\documentclass{article}

\usepackage{microtype}
\usepackage{graphicx}
\usepackage{subfigure}
\usepackage{booktabs} 
\usepackage[dvipsnames]{xcolor}

\usepackage{colortbl}
\usepackage{multirow,array,tabularx, float}
\usepackage{amsmath}
\usepackage{afterpage}

\definecolor{Gray}{gray}{0.95}
\definecolor{DarkGray}{gray}{0.5}
\definecolor{LightCyan}{rgb}{0.88,1,1}
\definecolor{cornsilk}{rgb}{1.0, 0.97, 0.86}

\usepackage[preprint, nonatbib]{neurips_2023}

\usepackage[utf8]{inputenc} 
\usepackage[T1]{fontenc}    
\usepackage{hyperref}       
\usepackage{url}            
\usepackage{booktabs}       
\usepackage{amsfonts}       
\usepackage{nicefrac}       
\usepackage{microtype}      
\usepackage{xcolor}         

\title{PolyThrottle: Energy-efficient Neural Network Inference on Edge Devices}

\author{%
  Minghao Yan\\
  Department of Computer Sciences\\
  University of Wisconsin-Madison\\
  \texttt{myan@cs.wisc.edu} \\
  \And
  Hongyi Wang\\
  School of Computer Science\\
  Carnegie Mellon University\\
  \texttt{hongyiwa@andrew.cmu.edu} \\
  \AND
  Shivaram Venkataraman\\
  Department of Computer Sciences\\
  University of Wisconsin-Madison\\
  \texttt{shivaram@cs.wisc.edu} \\
}

\begin{document}

\maketitle

\begin{abstract}
As neural networks (NN) are deployed across diverse sectors, their energy demand correspondingly grows. While several prior works have focused on reducing energy consumption during training, the continuous operation of ML-powered systems leads to significant energy use during inference. This paper investigates how the configuration of on-device hardware—elements such as GPU, memory, and CPU frequency, often neglected in prior studies, affects energy consumption for NN inference with regular fine-tuning. We propose PolyThrottle, a solution that optimizes configurations across individual hardware components using Constrained Bayesian Optimization in an energy-conserving manner. Our empirical evaluation uncovers novel facets of the energy-performance equilibrium showing that we can save up to 36 percent of energy for popular models. We also validate that PolyThrottle can quickly converge towards near-optimal settings while satisfying application constraints.
\end{abstract}

\section{Introduction}
The rapid advancements in neural networks and their deployment across various industries have revolutionized multiple aspects of our lives. However, this sophisticated technology carries a drawback: high energy consumption which poses serious sustainability and environmental challenges~\cite{anderson2022treehouse, gupta2022chasing, cao2020energy, wolff2020carbontracker}. Emerging applications such as autonomous driving systems and smart home assistants require real-time decision-making capabilities~\cite{jetson-solution}, and as we integrate NNs into an ever-growing number of devices, their collective energy footprint poses a considerable burden to our environment~\cite{wu2022sustainable, schwartz2020green, lacoste2019quantifying}. Moreover, considering that many devices operate on battery power, curbing energy consumption not only alleviates environmental concerns but also prolongs battery life, making low-energy NN models highly desirable for numerous use cases. 

In prior literature, strategies for reducing energy consumption revolve around designing more efficient neural network architectures~\cite{howard2017mobilenets, tan2019efficientnet}, quantization~\cite{kim2021bert, banner2018scalable, courbariaux2015binaryconnect, courbariaux2014training, gholami2021survey}, or optimizing maximum GPU frequency~\cite{you2022zeus, gu2023energy}. From our experiments, we make new observations about the tradeoffs between energy consumption, inference latency, and various other hardware configurations. Memory frequency, for example, emerges as a significant contributor to energy consumption (as shown in Figure~\ref{fig:3d_vis_tx2}), beyond the commonly investigated relationship between maximum GPU compute frequency and energy consumption. Table~\ref{tab:memTX2} shows that even with optimal maximum GPU frequency, we can save up to $25\%$ energy by further tuning memory frequency. In addition, minimum GPU frequency also proves to be of importance in certain cases, as shown in Figure~\ref{fig:heatmap_orin} and Table~\ref{tab:minGPU}. 

\begin{figure*}
\centering
\includegraphics[width=0.8\linewidth]{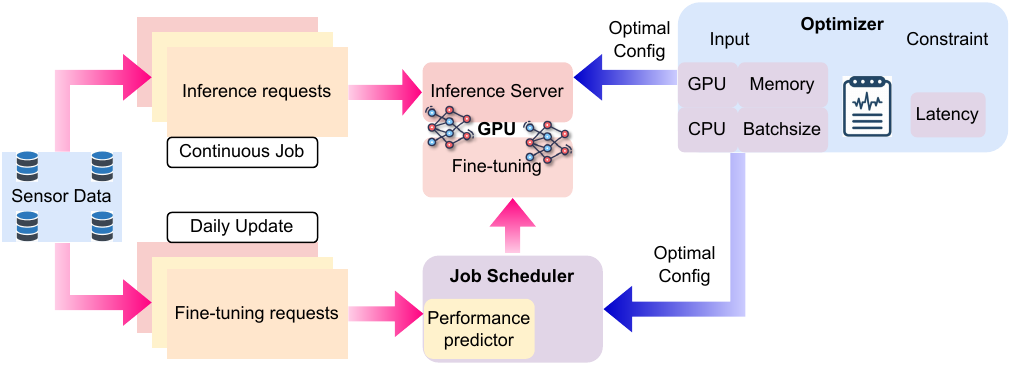}
\caption{Figure illustrating the overall workflow of PolyThrottle. The optimizer first identifies the optimal hardware configuration for a given model. When new data arrives, the inference server handles the inference requests. Upon receiving a fine-tuning request, our performance predictor estimates whether time-sharing inference and fine-tuning workloads would result in SLO violations. Then the predictor searches for feasible adjustments to meet the SLO constraints. If such adjustments are identified, the system implements the changes and schedules fine-tuning requests until completion.}
\label{fig:workflow}
\end{figure*}

\begin{figure*}
\centering
\includegraphics[width=\linewidth]{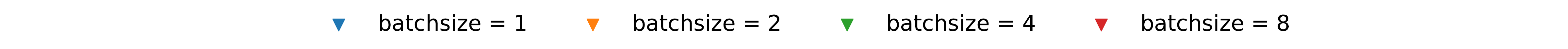}
\includegraphics[width=0.4\linewidth]{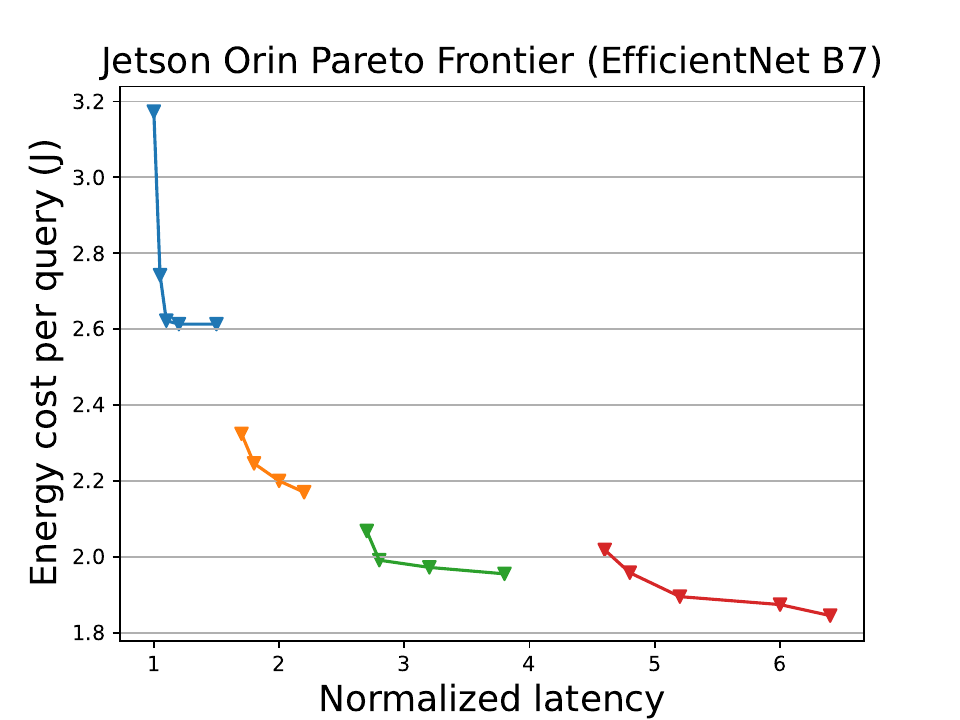}
\includegraphics[width=0.4\linewidth]{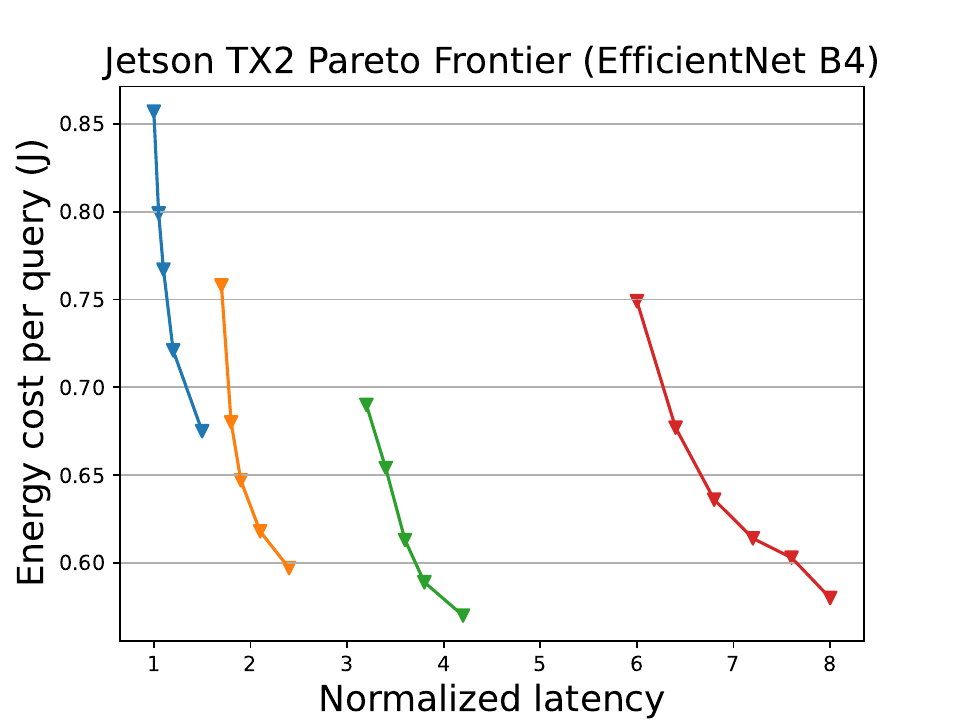}
\caption{\textbf{Left} figure shows the Pareto Frontier of energy vs. latency tradeoff for various batch sizes on EfficientNet B7 on Jetson Orin. \textbf{Right} figure shows the Pareto Frontier of energy vs. latency tradeoff for various batch sizes on EfficientNet B4 on Jetson TX2. Each data point in this plot is representative of a unique hardware configuration, and each line corresponds to a batch size. The figure shows that the tradeoff does not always conform to the same pattern across varied hardware platforms and models.}
\label{fig:pareto}
\end{figure*}

We also observe that a simple linear relationship falls short of capturing the tradeoff between energy consumption, neural network inference latency, and hardware configurations. The complexity of this tradeoff is illustrated by the Pareto Frontier in Figure ~\ref{fig:pareto}. This nuanced interplay between energy consumption and latency poses a challenging question: \textit{How can we find a near-optimal configuration that closely aligns with this boundary?}

Designing an efficient framework to answer the above question is challenging due to the large configuration space, the need to re-tune each model and hardware, and frequent fine-tuning operations. A naive approach, such as grid search, is inefficient and can take hours to find the optimal solution for a given model and desired batch size on a given hardware. The uncertainty in inference latency, especially at smaller batch sizes~\cite{gujarati2020serving}, further exacerbates the challenge. Furthermore, given that distinct hardware platforms and NN models display unique energy consumption patterns (Section~\ref{sec:opportunities}), relying on a universally applicable pre-computed optimal configuration is not feasible. Every deployed device must be equipped to quickly identify its best configuration tailored to its specific workload. Finally, in production environments, daily fine-tuning is often necessary to adapt to a dynamic external environment and integrate new data~\cite{cai2019device, cai2020tinytl}. This demands a mechanism that can quickly adjust configurations to complete fine-tuning requests in time while ensuring the online inference workloads meet Service Level Objectives (SLOs). 

In this paper, we explore the interplay between inference latency, energy consumption, and hardware frequency and propose PolyThrottle as our solution. PolyThrottle takes a holistic approach, optimizing various hardware components and batch sizes concurrently to identify near-optimal hardware configurations under a predefined latency SLO. PolyThrottle complements existing efforts to reduce inference latency, including pruning, quantization, and knowledge distillation. We use Constrained Bayes Optimization with GPU, memory, CPU frequencies, and batch size as features, and latency SLO as a constraint to design an efficient framework that automatically adjusts configurations, enabling convergence towards near-optimal settings. Furthermore, PolyThrottle uses a performance prediction model to schedule fine-tuning operations without disrupting ongoing online inference requests. We integrate PolyThrottle into Nvidia Triton on Jetson TX2 and Orin and evaluate on state-of-the-art CV and NLP models, including EfficientNet and Bert~\cite{tan2019efficientnet, devlin2018bert}.

To summarize, our key contributions include:

1. We examine the influence of hardware components beyond GPUs on energy consumption, delineate new tradeoffs between energy consumption and inference performance, and reveal new possibilities for optimization.

2. We construct an adaptive framework that efficiently finds energy-optimal hardware configurations. To accomplish this, we employ Constrained Bayesian Optimization.

3. We develop a performance model to capture the interaction between inference and fine-tuning processes. We use this model to schedule fine-tuning requests and carry out real-time modifications to meet inference SLOs.

4. We implement and evaluate PolyThrottle on a state-of-the-art inference server on Jetson TX2 and Orin. With minimal overheads, PolyThrottle reduces energy consumption per query by up to $36\%$.

\section{Motivation} \label{sec:motivation}
Many deep neural networks have been deployed on edge devices to perform tasks such as image classification, object detection, and dialogue systems. Scenarios including smart home assistants~\cite{he2020fedml}, inventory and supply chain monitoring~\cite{jetson-solution}, and autopilot~\cite{gog2022d3} often use battery-based devices that contain GPUs to perform the aforementioned tasks. In these scenarios, pre-trained models are installed on the devices where the inference workload is deployed. 

Prior works have focused on optimizing the energy consumption of GPUs~\cite{wang2020benchmarking, wang2021dynamic, tang2019impact, strubell2019energy, mei2017survey} in cloud scenarios~\cite{qiao2021pollux, wan2020alert, hodak2019power} and training settings~\cite{wang2020blink, peng2019generic, kang2022cost}.  On-device inference workloads exhibit different characteristics and warrant separate attention. In this section, we outline previous efforts in optimizing on-device neural network inference and discuss our approach to holistically optimize energy consumption. 

\subsection{On-device Neural Network Deployment}
Prior work in optimizing on-device neural network inference focuses on quantization~\cite{kim2021bert, banner2018scalable, courbariaux2015binaryconnect, courbariaux2014training, gholami2021survey}, designing hardware-friendly network architectures~\cite{first2019xu, lee2019device, sanh2019distilbert, touvron2021training, howard2019searching}, and leveraging hardware components specific to mobile settings, such as DSPs~\cite{lane2015can}. Our work explores an orthogonal dimension and aims to answer a different question: \textbf{Given a neural network to deploy on a specific device, how can we tune the device to reduce energy consumption? }

In our work, we focus on edge devices that contain CPUs, memory, and GPUs. These devices are generally more powerful than DSPs often found on mobile devices. One such example is the Nvidia Jetson series, which is capable of handling a wide array of applications, ranging from AI to robotics and embedded IoT solutions~\cite{jetson-solution}. The devices also come with dynamic voltage and frequency scaling (DVFS) capabilities that allow for the optimization of power consumption and thermal management during complex computational tasks. The Jetson series features a unified memory shared by both CPU and GPUs. We refer to the operating frequency of CPU, GPU, and shared memory as CPU frequency, GPU frequency, and memory frequency in this paper.

\textbf{Case Study on Inventory Management:} 
To understand the system requirements in edge NN inference, we next describe a case study of how NNs are deployed in an inventory management company. From our conversations, Company A works with Customer B to deploy neural networks on edge devices to optimize inventory management. To comply with regulations and protect privacy, data from each inventory site are required to be stored locally. The vast difference in the layout of the inventories makes it impossible to pre-train the model on data from every warehouse. Therefore, these devices come with a pre-trained model based on data from a small sample of inventories, which may have significantly different layouts and external environments compared to the actual deployment venue. Consequently, daily fine-tuning is required to enhance performance in the deployed sites, as the environment continually evolves. Similar arguments apply to smart home devices, where a model is pre-trained on selected properties, but the deployed households may be much more diverse. To address privacy concerns, on-device fine-tuning of neural networks is preferred, as it keeps sensitive data locally. Therefore, edge devices often need to run both inference and periodic fine-tuning. Combining multiple workloads on edge devices can lead to SLO violations due to interference and increased energy use.

\subsection{Holistic Energy Consumption Optimization}
Some recent works have explored reducing energy consumption by optimizing for batch size and GPU maximum frequency~\cite{you2022zeus, nabavinejad2021batchsizer, komoda2013power, gu2023energy} and developing power models for modern GPUs~\cite{kandiah2021accelwattch, hong2010integrated, arafa2020verified, lowe2020gem5}. In this work, we argue that other hardware components also cause energy inefficiency and require separate optimization. We perform a grid search over GPU, memory, and CPU frequencies and various batch sizes to examine the Pareto frontier of inference latency and energy consumption. Figure~\ref{fig:pareto} shows the tradeoff between the per-query energy consumption and inference latency (normalized to the optimal latency) on Jetson TX2 and Jetson Orin. Each point in the figure represents the optimal configuration that we find through grid search under a given inference latency budget and batch size.  As Figure \ref{fig:pareto} shows, the Pareto frontier is not smooth globally and is difficult to capture by a simple model, which warrants more sophisticated optimization techniques to quickly converge to a hardware configuration that lies on the Pareto Frontier~\cite{censor1977pareto}.

Zeus~\cite{you2022zeus} attempts to reduce the energy consumption of neural network training by changing the GPU power limit and tuning training batch size. PolyThrottle also includes these two factors. In Zeus~\cite{you2022zeus}, the focus is on training workloads in data center settings, where batch size tuning helps achieve an accuracy threshold in an energy-efficient way. We include batch size as part of PolyThrottle as it provides a trade-off between inference latency and throughput. Our empirical evaluation reveals new avenues available for optimization which complicates the search space, as we describe next.

\section{Opportunities} \label{sec:opportunities}
In this section, we perform empirical experiments to uncover new opportunities for optimizing energy use in NN inference. As discussed in Section~\ref{sec:motivation}, prior work did not study how memory frequency, minimum GPU frequency, and CPU frequency play a role in energy consumption. This is partially limited by hardware constraints. Specialized power rails need to be built into the device during manufacturing to enable accurate measurement of energy consumption associated with each component. We leverage two Jetson developer kits, TX2 and Orin, which offer native support for component-wise energy consumption measurement and frequency tuning, to study how these frequencies impact inference latency and energy consumption in modern deep learning workloads. We find that the default frequencies are much larger than optimal and throttling all these frequency knobs offer energy consumption reduction with minimal impact on inference latency. 

\begin{figure}
\centering
\includegraphics[width=0.4\linewidth]{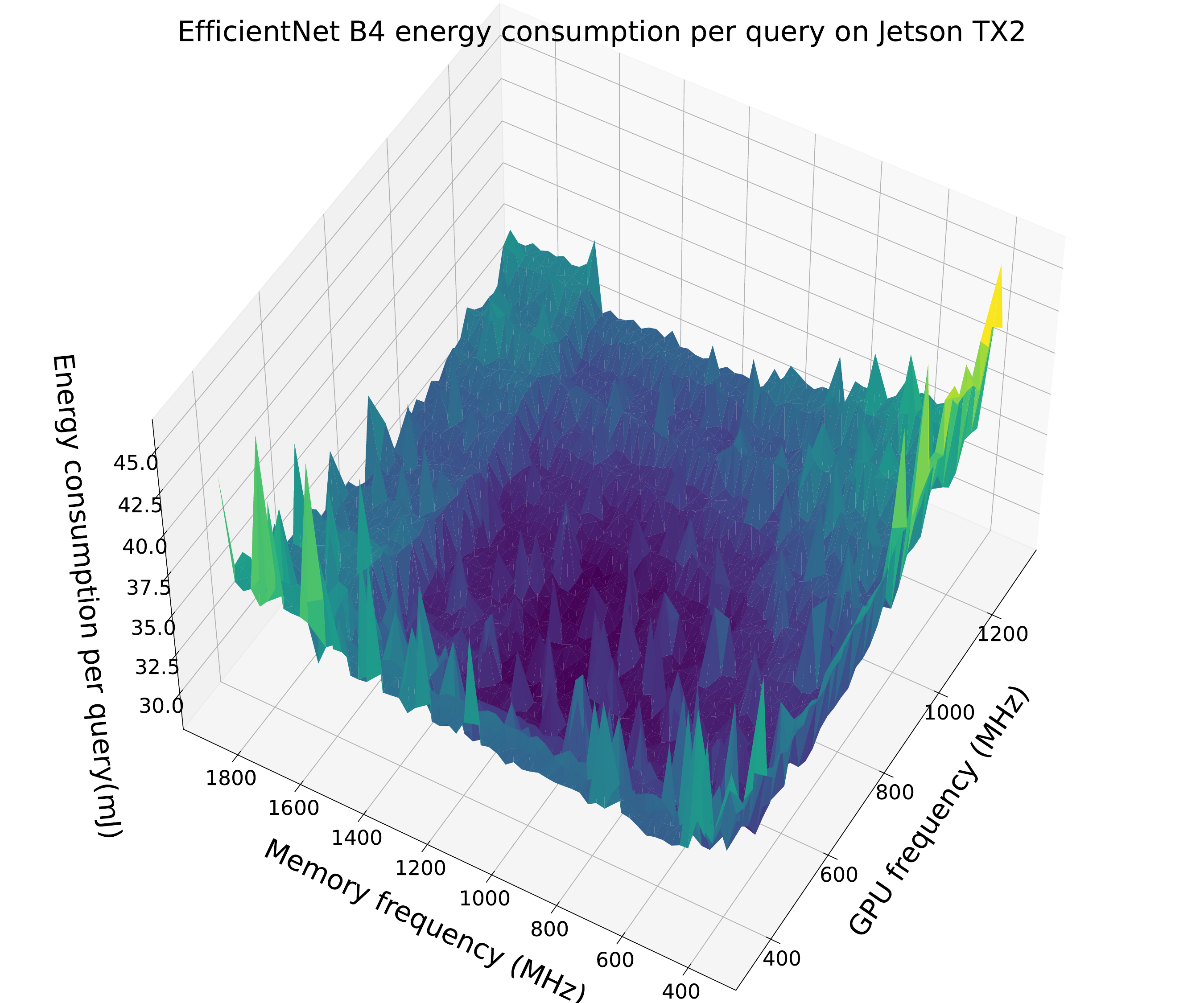}
\includegraphics[width=0.4\linewidth]{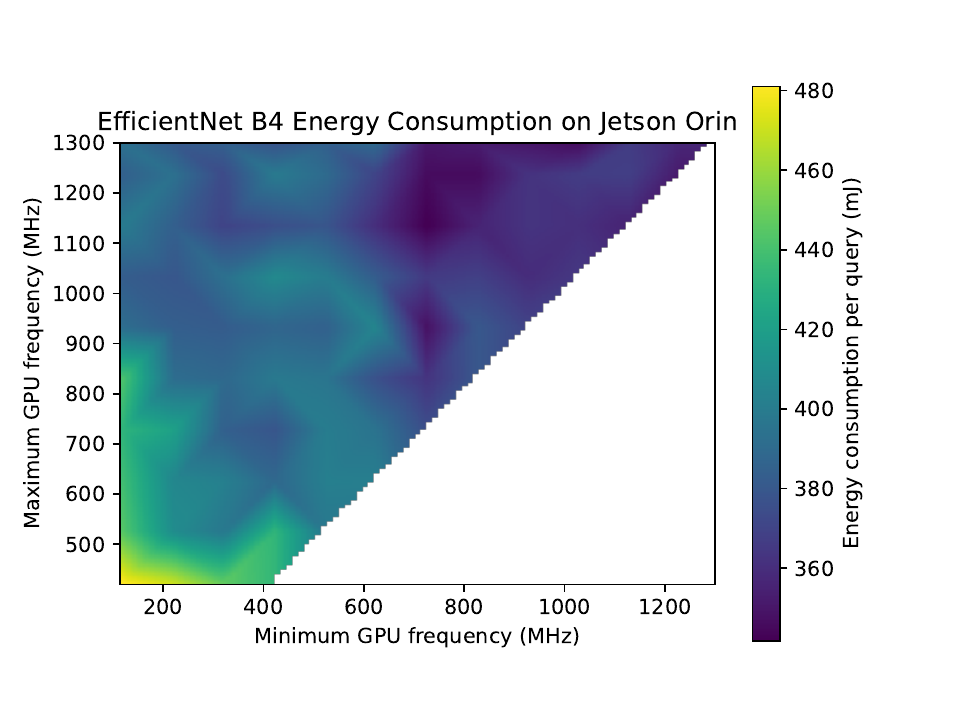}
\caption{Left figure shows per query energy cost as we vary the GPU frequency and memory frequency for EfficientNet B4 on Jetson TX2 versus varying memory and GPU frequency with batch size fixed at 1. Right figure shows per query energy cost as we vary the minimum and maximum GPU frequency. As we increase the minimum GPU frequency, energy cost decreases.}
\label{fig:heatmap_orin}

\label{fig:3d_vis_tx2}
\end{figure}

Figure~\ref{fig:3d_vis_tx2} illustrates the energy optimization landscape when varying GPU and memory frequencies, without imposing any constraints on latency SLO. The plot reveals that, without any other constraints, the energy optimization landscape generally exhibits a bowl shape. However, this shape varies depending on the models, devices, and other hyperparameters, such as batch sizes (See Appendix \ref{appen:exp} for more results). Next, we dive into how each hardware component affects inference energy consumption.

\textbf{Setup: } Experiments in this section are performed with 16-bit floating point number precision, as it has been demonstrated to have minimal impact on model accuracy in practice. We use Bert and EfficientNet models and vary the EfficientNet model size between B0, B4, B7 (Table~\ref{tab:enet_scaling}). 

\textbf{Memory frequency experiment:} For each model, we fix the GPU frequency at the optimal frequency determined by grid-search of all possible frequency configurations. We then examine the tradeoff between inference latency and energy consumption as we progressively throttle memory frequency. The range of available memory frequencies can be found in Table ~\ref{tab:range}. 

\textbf{Results: } Table~\ref{tab:memTX2} reveals that memory frequency plays a vital role in reducing energy consumption. The savings provided by memory frequency tuning are similar and consistent across models on both hardware platforms, ranging from approximately $12\%$ to $25\%$. This indicates that the default memory frequency is higher than optimal for modern Deep Learning workloads. For heavy workloads such as Bert, memory tuning can account for the majority of the energy consumption reduction. 
This can be partially attributed to the memory-bound nature of Transformer-based models~\cite{ivanov2021data}. Our result demonstrates that systems that aim to optimize energy use in neural network inference need to take memory frequency into account.

\begin{table}[t]
 \centering
 \caption{This table shows the energy frequency range for Jetson TX2 and Jetson Nano in MHz.}
\begin{tabular}{ lcccc } 

 \toprule
 \rowcolor{Gray} Device & Min GPU & Max GPU & Min Mem & Max Mem\\
 \midrule
 
 TX2 & 114.75 & 1300.5 & 40.8 & 1866\\
 Orin & 114.75 & 1300.5 & 204 & 3199\\  
  \bottomrule
 \end{tabular}
 \label{tab:range}
\end{table}

\begin{table}[t]
 \centering
 \caption{This table shows the optimal memory frequency (in MHz) and the corresponding energy savings for various models on Jetson TX2. B0/B4/B7 represent different models in the Efficient Net series.}
\begin{tabular}{ lcc } 

 \toprule
 \rowcolor{Gray} Model & Energy Reduction & Optimal Mem Freq\\
 \midrule
 
 B0 & 14.9\% & 1331\\ 
 B4 & 14.3\% & 1331\\ 
 B7 & 12.0\% & 1331\\
 Bert Base & 25.4\% & 1062\\  
  \bottomrule
 \end{tabular}
 \label{tab:memTX2}
\end{table}

\textbf{CPU Frequency Experiment: } CPUs are only used for data pre-processing. Thus, we first measure the time spent in the data processing part of the inference pipeline. Next, we measure the energy saved by throttling the CPU frequency and assess the inference latency slowdown caused by reducing CPU frequency. The data preprocessing we perform is standard in almost all image processing and object detection pipelines, where we read the raw image file, convert it to an RGB scale, resize it, and reorient it to the desired input resolution and data layout.

\textbf{Results: } The preprocessing time across different EfficientNet models remains constant since the operations performed are identical. As a result, the relative impact of CPU tuning on overall energy consumption depends on the ratio between preprocessing time and inference time. As the model size increases and inference duration increases, the influence of CPU tuning on overall energy consumption decreases. We observe that on both Jetson TX2 and Orin platforms, CPU tuning can decrease preprocessing energy consumption by approximately $30\%$. Depending on the model, quantization level, and batch size, this results in up to a $6\%$ reduction in overall energy consumption.

\begin{table}[t]
 \centering
 \caption{This table shows the optimal minimum GPU frequency (MHz) and the corresponding energy savings for various models with 16-bit floating-point precision on Jetson Orin. B0/B4/B7 represent different models in the Efficient Net series.}
\begin{tabular}{ lccc } 

 \toprule
\rowcolor{Gray}  & &  & Optimal Min \\
\rowcolor{Gray} \multirow{-2}{*}{{Model}} & \multirow{-2}{*}{{Size}}& \multirow{-2}{*}{{Energy Reduction}} & GPU Freq\\
 \midrule
 B0 & 5.3M & 47.6\% & 1236\\ 
 B4 & 19M & 29.1\% & 1033\\ 
 B7 & 66M & 8.8\% & 1134\\
 Bert Base & 110M & 1.1\% & 217\\  
  \bottomrule
 \end{tabular}
 \label{tab:minGPU}
\end{table}

\textbf{Minimum GPU frequency experiment:} We maintain the default hardware configuration and only adjust the minimum GPU frequency on Jetson Orin. Increasing the minimum GPU frequency forces the GPU DVFS mechanism to operate within a smaller range.  We scale the model from EfficientNet B0 to EfficientNet B7 to illustrate the effect of the GPU minimum frequency on inference latency.

\textbf{Results: } Table~\ref{tab:minGPU} indicates that tuning the minimum GPU frequency can significantly reduce energy consumption when the workload cannot fully utilize the computational power of the hardware. Notably, energy consumption and inference latency are reduced by forcing the GPU to operate at a higher frequency. This differs from the tradeoff observed in other experiments, where we exchange inference latency for lower energy consumption. Tuning minimum GPU frequency can nearly halve the energy consumption for small models. As computational power becomes saturated with increasing model size, the return on tuning the minimum GPU frequency diminishes. 

Figure~\ref{fig:heatmap_orin} shows the per query energy cost as we vary the minimum and maximum GPU frequency. It shows that increasing the minimum GPU frequency from the default minimum leads to lower energy costs and inference latency. 

\section{Architecture Overview}

To take advantage of the opportunities described in the previous section, we design PolyThrottle, a system that navigates the tradeoff between latency SLO, batch size, and energy. PolyThrottle optimizes for the most energy-efficient hardware configurations under performance constraints and handles scheduling of on-device fine-tuning.

Figure~\ref{fig:workflow} shows a high-level overview of PolyThrottle's workflow. In a production environment, sensors on the edge devices continuously collect data and send the data to the deployed model for inference. In the meantime, to adapt to a changing environment and data patterns, these data are also saved for fine-tuning later. Due to the limited computation resources on these edge devices, fine-tuning workloads are often scheduled in conjunction with the continuously running inference requests. To address the challenges in model deployment on edge devices, PolyThrottle consists of two key components: 

1. An optimization framework that finds optimal hardware configurations for a given model under predetermined SLOs using few samples.

2. A performance predictor and scheduler to dynamically schedule fine-tuning requests and adjust for the optimal hardware configuration while satisfying SLO.

PolyThrottle tackles these challenges separately. Offline, we automatically find the best CPU frequency, GPU frequency, memory frequency, and recommended batch size for inference requests that satisfy the latency constraints while minimizing per-query energy consumption. We discuss the details of the optimization procedure in Section~\ref{sec:tuning}. We also show that our formulation can find near-optimal energy configurations in a few minutes using just a handful of samples. Compared to the lifespan of long-running inference workloads, the overhead is negligible. 

The optimal configuration is then installed on the inference server. At runtime, the client program processes the input and sends inference requests to the inference server. Meanwhile, if there are pending fine-tuning requests, the performance predictor predicts the inference latency when running concurrent fine-tuning, and decides whether it is possible to satisfy the latency SLO if fine-tuning is scheduled concurrently. A detailed discussion on performance prediction can be found in Section~\ref{sec:perf-pred}. The scheduler then decides what the new configuration that can satisfy the latency SLO while minimizing per-query energy consumption is. If such a configuration is attainable, it will schedule fine-tuning requests iteration-by-iteration until all pending requests are finished.

\noindent\textbf{Online vs. Offline:} Adjusting the frequency of each hardware component entails writing to one or multiple hardware configuration files, a process that takes approximately 17ms each. On Jetson TX2 and Orin, each CPU core, GPU, and memory has a separate configuration file that determines operating frequency. As a result, setting the operating frequencies for CPUs, GPU, and memory could require up to 150ms. This duration could exceed the latency SLO for many applications, and this is without accounting for the additional overhead of completing frequency changes. Since the latency SLO for a specific workload does not change frequently, PolyThrottle determines the optimal hardware configuration before deployment and only performs online adjustments to accommodate fine-tuning workloads.

\section{Problem Formulation: Two-phase Tuning} \label{sec:tuning}
Our objective is to automatically find the optimal hardware configurations that minimize energy consumption while satisfying latency SLOs. Formally, we are solving the optimization problem:
\begin{align*}
 \min \quad &~ f(x_{CPU}, \  x_{GPU_{min}}, \  x_{GPU_{max}}, \ x_{Mem}, \  b) \\
 \mathrm{s.t.} \quad &~ t(x_{CPU}, \  x_{GPU_{min}}, \  x_{GPU_{max}}, \  x_{Mem}, \  b) \leq c 
\end{align*}
where $f$ and $t$ represent the energy consumption and latency associated with a workload under the given hardware configurations and batch size. We use $x_{CPU}, \  x_{GPU_{min}}, \  x_{GPU_{max}},  x_{Mem}$ to denote the frequency limit of CPU, GPU, and memory on the device, and use $b$ to denote the maximum batch size. We use $c$ to denote the inference latency SLO limit, which is set by application users. This optimization problem is challenging on several fronts: 

1. The search space is large, and performing a grid search will take hours depending on the model size. On TX2 and Orin, there are 5005 and 1820 points in the grid if we allow 5 different batch sizes. An exhaustive search would take 14 and 5 hours, respectively.

2. To satisfy latency constraints, it is hard to decouple each dimension and optimize them separately as they jointly affect inference latency in a non-trivial way.

3. The optimization landscape may vary across models and devices.

We observe that CPU frequency can be decoupled from GPU frequency, memory frequency, and batch size, as it mainly affects the preprocessing latency and energy consumption. We pipeline the requests so different requests can use CPU and GPU resources at the same time to increase inference throughput.

Based on this observation, we propose a two-phase hardware tuning framework, where CPU tuning is done separately from tuning other hardware components. The challenge that remains is to efficiently optimize for an unknown function with noise. As shown in Figure~\ref{fig:3d_vis_tx2} and \ref{fig:heatmap_orin}, the performance of neural network inference with changes in memory and GPU frequency is difficult to predict, therefore, a good solution must be able to handle the variance while converging to a near-optimal configuration in a sample-efficient fashion. This requires the method to adaptively balance the tradeoff between exploration and exploitation. To solve this, we formulate the optimization problem as a Bayesian Optimization problem and leverage recent advances in the field to incorporate the SLO constraints unique to our setting.

\subsection{Constrained Bayesian Optimization}
Bayesian Optimization is a prevalent method for hyperparameter tuning~\cite{kandasamy2020tuning, klein2015towards}, as it can optimize various black-box functions. This method is especially advantageous when evaluating the objective function is expensive and requires a substantial amount of time and resources.

However, some applications may involve constraints that must be satisfied in addition to optimizing the objective function. Constrained Bayesian Optimization (CBO)~\cite{gardner14} is an extension of Bayesian Optimization that tackles this challenge by incorporating constraints into the optimization process.

In CBO, the objective function and constraints are treated as distinct functions. The optimization algorithm seeks to identify the set of input parameters that maximize the objective function while adhering to the constraints. These constraints are usually expressed as inequality constraints that must be satisfied during the optimization process. The expected constrained improvement acquisition function in CBO is defined as follows: $EI_C(\hat{x}) = PF(\hat{x}) \times EI(\hat{x})$.

Here $EI(\hat{x})$ represents the expected improvement (EI) \cite{brochu2010tutorial} within an unconstrained Bayesian Optimization scenario, while $PF(\hat{x})$ is a univariate Gaussian cumulative distribution function, delineating the anticipated probability of whether $\hat{x}$ can fulfill the constraints. Intuitively, EI chooses the next configuration by optimizing the expected improvement relative to the best recently explored configuration. In PolyThrottle, we choose EI since our empirical findings and corroborations from additional studies \cite{alipourfard2017cherrypick} show that EI performs better than other widely-used acquisition functions \cite{snoek2012practical}.

CBO \cite{gardner14} also employs a joint prior distribution over the objective and constraint functions that captures their correlation structure. This joint prior is constructed by assuming that the objective and constraint functions are drawn from a multivariate Gaussian distribution with a parameterized mean vector and covariance matrix. These hyperparameters are learned from data using maximum likelihood estimation.

During the optimization process, the algorithm uses this joint prior to compute an acquisition function that balances exploration (sampling points with high uncertainty) and exploitation (sampling points where the objective function is expected to be low and subject to feasibility constraints). The algorithm then selects the next point to evaluate based on this acquisition function. During each iteration, the algorithm will test whether the selected configuration violates any of the given constraints and take the result into account for the next iteration. Encoding more system-specific hints as constraints can be of independent research interests, however, we show in Section~\ref{sec:exp} that the current formulation performs well under a variety of scenarios.

\section{Modeling Workload Interference} \label{sec:perf-pred}

Consider the case where we run an inference workload and aim to support fine-tuning without interfering with the online inference process. When a fine-tuning request arrives, we need to decide if it is possible to execute the fine-tuning request without violating inference SLOs. Time-sharing has been the default method for sharing GPU workloads. In time-sharing, shared workloads use different time slices and alternate GPU use between them. Recently, CUDA streams, Multiprocess Service (MPS)~\cite{nvidia-cuda-streams} and MIG~\cite{nvidia-mig} have been proposed to perform space-sharing on GPUs. However, these approaches are not supported on edge GPU devices~\cite{bai2020pipeswitch, zhao2023muxflow, yu2020fine, wu2021switchflow}. Given this setup, we propose building a performance model that can predict the inference latency in the presence of fine-tuning requests and only execute fine-tuning requests if the predicted latency can satisfy SLO.

\textbf{Feature selection: } To build the performance model, we leverage the following insights to select features: 1. In convolutional neural networks, the 2D convolution layers' performance largely determines the overall performance of the network. Its latency is correlated to the number of floating point operations (FLOPs) required during forward / backward propagation. 2. The ratio between the number of FLOPs and the number of memory accesses, also known as arithmetic intensity, together with total FLOPs, encapsulates whether a neural network is compute-bound or memory-bound. Using these insights, we add the following features to our model: Inference FLOPs, Inference Arithmetic Intensity, Fine-tuning FLOPs, Fine-tuning Arithmetic Intensity, and Batchsize.

\textbf{Model selection: } We propose using a linear model to predict inference latency when a fine-tuning workload is running concurrently on the same device. The model aims to capture how the proposed variables affect the resource contention between the inference workload and the fine-tuning workload, and therefore, affect the inference latency. The proposed model can be summarized as follows: 

\vspace{-0.5cm}
\begin{align*}
\text{Inference time} = &\theta_0 + \theta_1 \times \text{FLOPs}_{inf} + \theta_2 \times AI_{inf} \\
        & + \theta_3 \times \text{FLOPs}_{ft} + \theta_4 \times AI_{ft} \\
        & + \theta_5 \times \text{Batchsize}
\vspace{-0.3cm}
\end{align*}
\vspace{-0.5cm}

Given the above performance model, we use a Non-negative Least Squares (NNLS) solver to find the model that best fits the training data. An advantage of NNLS for linear models is that we can solve this with very few training data points~\cite{venkataraman2016ernest}. We collect a few samples on the provided model by varying the inference and fine-tuning batch sizes and the output dimension, which captures various fine-tuning settings. This model is used as part of the workload scheduler during deployment to predict whether it is possible to schedule a fine-tuning request. 

\textbf{Fine-tuning scheduler: } During inference, when there are outstanding fine-tuning requests, PolyThrottle uses the model to decide whether it is possible to schedule the request online without violating the SLO. When the model finds a feasible configuration, it adjusts accordingly until either all pending requests are finished or a new latency constraint is imposed.

\vspace{-0.3cm}

\section{Experiments} \label{sec:exp}

\begin{table}[t]
 \centering
 \caption{This table shows the scaling pattern of the EfficientNet model family.}
\begin{tabular}{ lccc } 

 \toprule
 \rowcolor{Gray} Model & Input dim / width & Width coef & Depth coef\\
 \midrule
B0 & 224 $\times$ 224 & 1.0 & 1.0 \\ 
B4 & 380 $\times$ 380 & 1.4 & 1.8  \\
B7 & 600 $\times$ 600 & 2.0 & 3.1 \\ 
  \bottomrule
 \end{tabular}
 \label{tab:enet_scaling}
\end{table}

\subsection{Setup}
\textbf{Hardware Platform:} Our experiments are conducted on the Jetson TX2 Developer Kit and Jetson Orin Developer Kit. 
To assess the energy consumption of our program, we employ the built-in power monitors on the Jetson TX2 and Jetson Orin Developer Kits. We also cross-validate our measurements with an external digital multimeter (See Appendix~\ref{appen:measurement} for more details on hardware and energy measurement).

\textbf{Workload Selection:} We base our experiments on the EfficientNet family and Bert models~\cite{tan2019efficientnet, devlin2018bert}. EfficientNet is chosen not only for its status as a state-of-the-art convolutional network in on-device and mobile settings but also for its principled approach to scaling the width, depth, and resolution of convolution layers. Table~\ref{tab:enet_scaling} summarizes the scaling pattern of EfficientNet from the smallest B0 to the largest B7. We select Bert to investigate energy usage patterns in a Transformer-based model~\cite{wolf2020transformers}, where the workload is more memory-bounded compared to convolution-based neural networks. Bert and its variants \cite{kim2021bert, devlin2018bert, sanh2019distilbert, tambe2021edgebert} are widely used for Question and Answering tasks \cite{rajpurkar2016squad}, making it applicable for numerous edge devices, such as smart home assistants and smart speakers. 

\textbf{Dataset: } We evaluate PolyThrottle on real-world traffic streams data \cite{shen2019nexus} and sample frames uniformly to feed into EfficientNet. For Bert, we evaluate on SQuAD \cite{rajpurkar2016squad} for Question Answering. Note that datasets do not affect PolyThrottle's performance since inference latency would not change significantly across datasets once the model is chosen.

\textbf{Implementation: } PolyThrottle is built on the Nvidia Triton inference server. To maximize performance, we generate TensorRT kernels that profile various data layouts and tiling strategies to identify the fastest execution graph for a given hardware platform. Our modules include a Bayesian optimizer for determining the best configuration, an inference client responsible for preprocessing and submitting requests to the inference server, and a performance predictor module integrated into the inference client for scheduling fine-tuning requests. We maintain separate queues for inference and fine-tuning requests.

\subsection{Efficiently Searching for Optimal Configuration} \label{sec:pareto}
In this experiment, we carry out an extensive empirical analysis of tuning various models across different hardware configurations while also adjusting the quantization level. We perform a grid search on EfficientNet B0, B4, B7, and Bert Base to examine the potential energy savings and identify the optimal GPU and memory frequencies for each model. We also adjust the quantization level for each tested model. We evaluate 16-bit and 32-bit floating point (FP16/FP32) precision. The optimal energy consumption and configuration referenced later in this section use the results obtained here as the baseline and optimal solution. Having obtained the optimal frequency using grid search, we next evaluate the average number of attempts it takes for PolyThrottle to find a solution within $5\%$ of the optimal solution. We compare our Constrained Bayesian Optimization (CBO) formulation against Random Search (RS).

\renewcommand{\floatpagefraction}{0.9}
\begin{figure*}
\centering
\includegraphics[width=\linewidth]{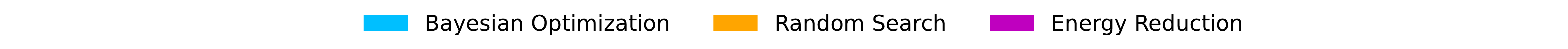}
\includegraphics[width=0.45\linewidth]{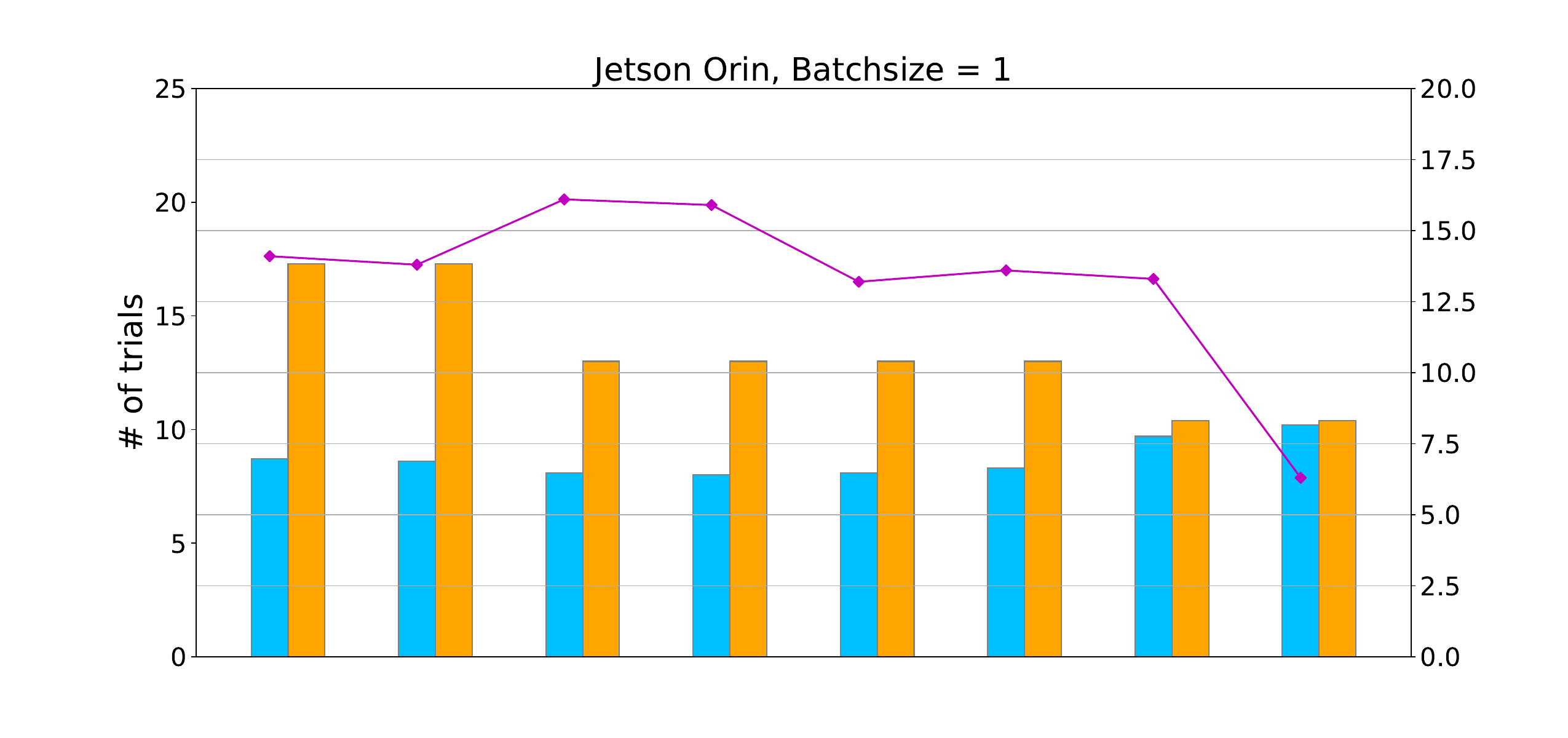}
\includegraphics[width=0.45\linewidth]{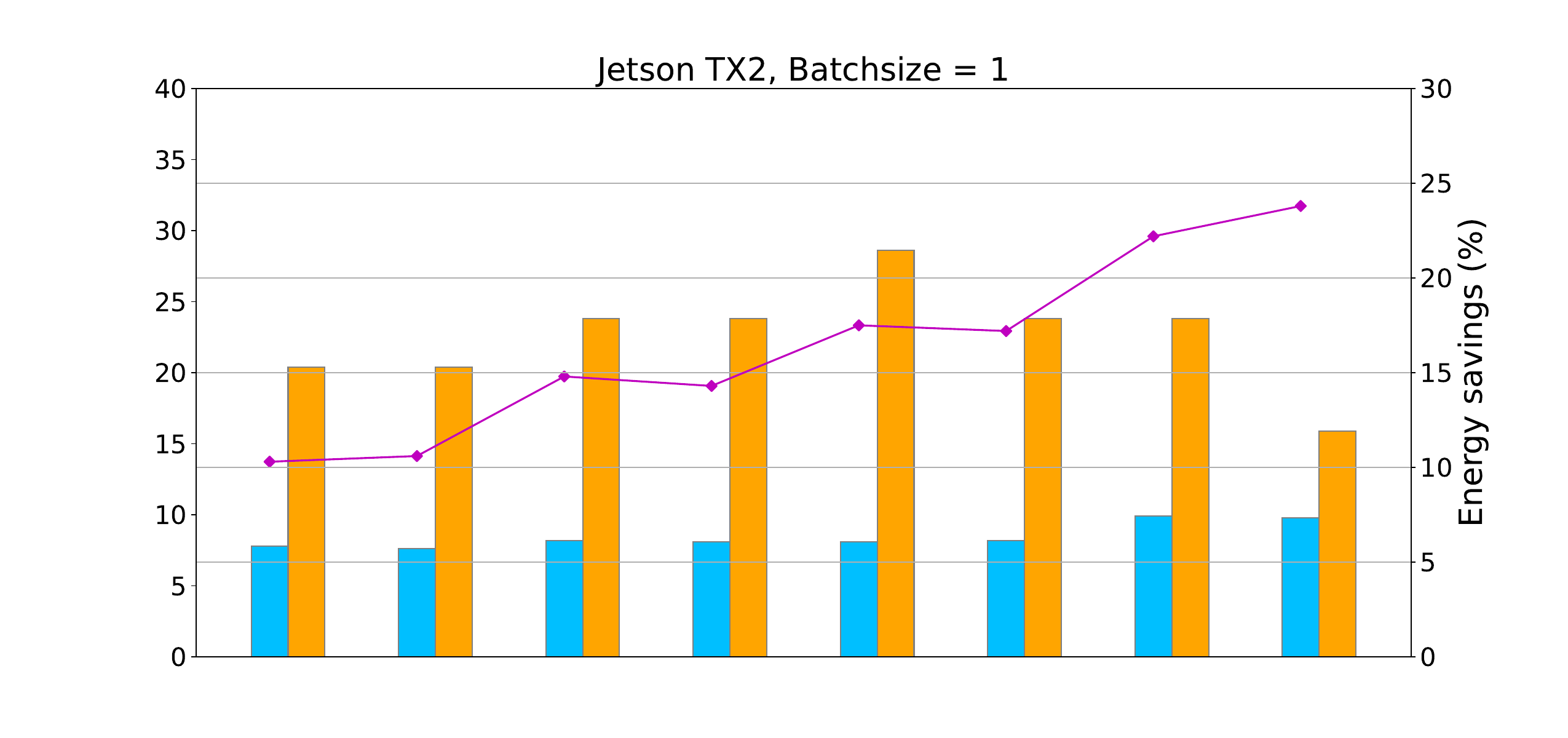}
\includegraphics[width=0.45\linewidth]{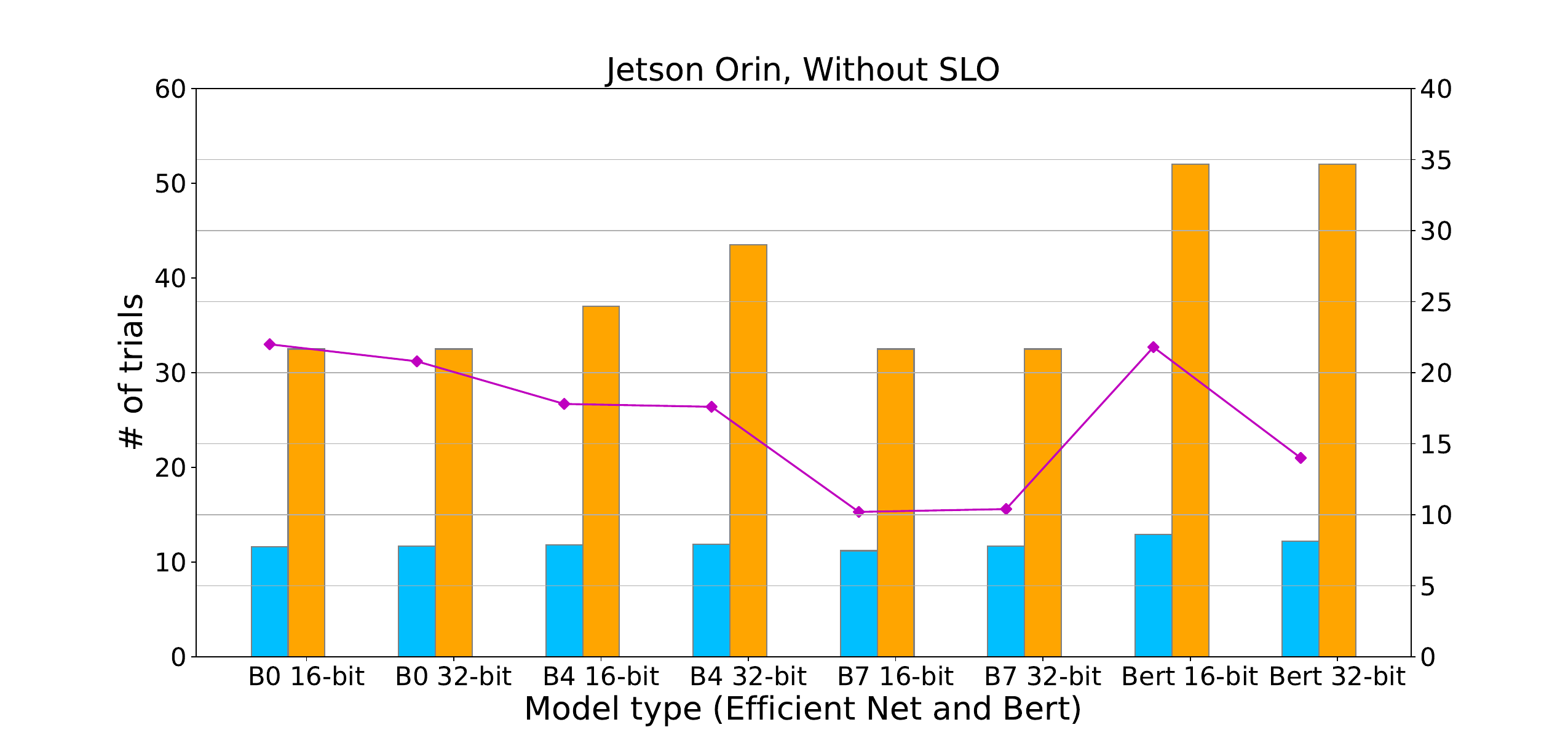}
\includegraphics[width=0.45\linewidth]{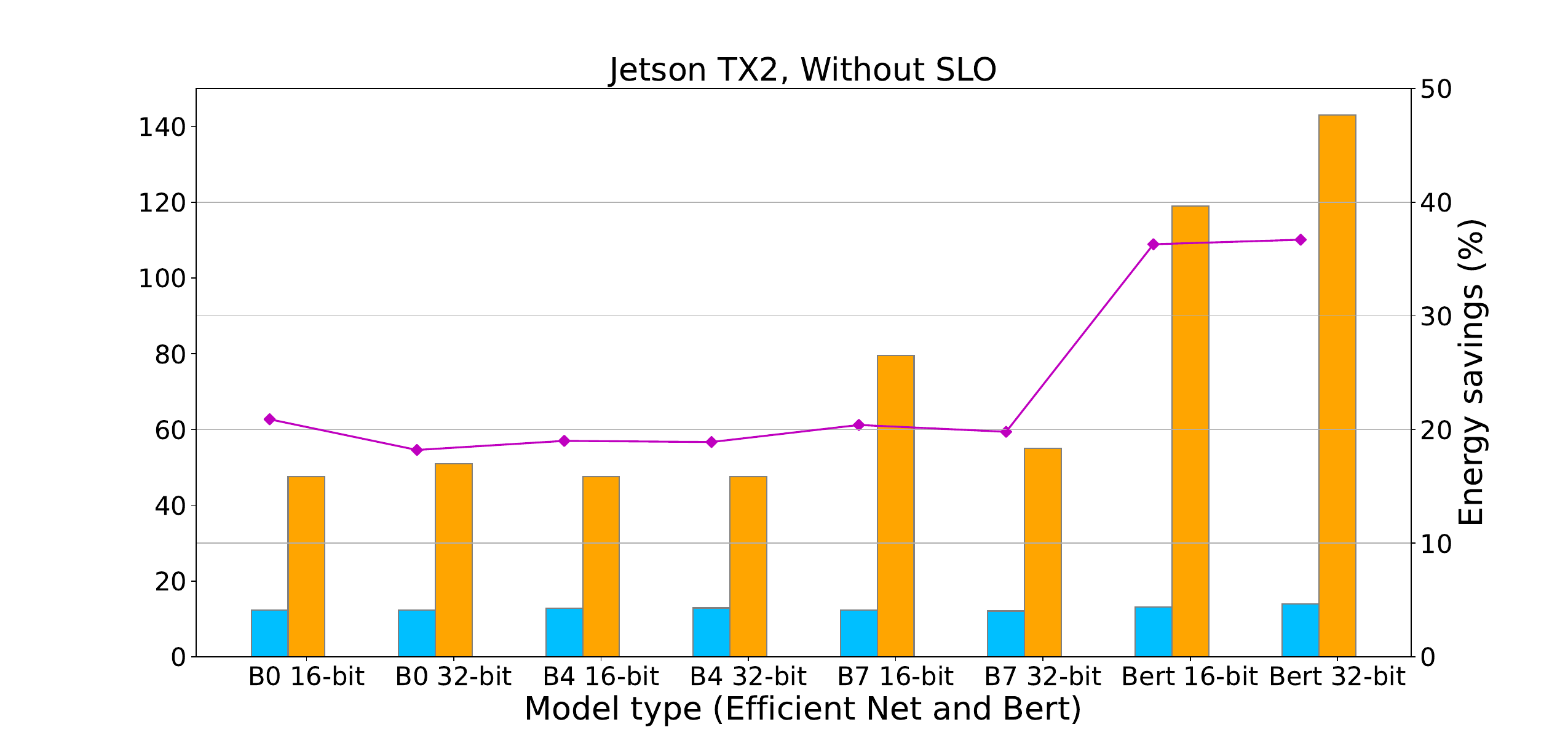}
\caption{This figure compares search efficiency between Constrained Bayesian Optimization and Random Search. The y-axis represents the number of attempts it takes to find a near-optimal configuration and the x-axis represents the deployed and associated quantization level. The \textbf{first row} corresponds to the setting where we set a latency target but restrict the batch size to 1. The \textbf{second row} where we relax the latency constraint and allow batching inference requests. }
\label{fig:bo_vs_rs}
\end{figure*}

\textbf{Experiment Settings: } We measure the average number of attempts needed to find a near-optimal configuration. For Random Search, we calculate the expected number of trials needed to find a near-optimal configuration based on the grid size by computing the fraction of near-optimal configurations and taking the reciprocal. For CBO, we set the $\xi$ parameter associated with the Expected Improvement function to 0.1 and initial random samples to be 5, which we find to work well across different models and hardware platforms. We conduct two experiments where we set different inference latency constraints, the results can be found in Figure \ref{fig:bo_vs_rs}:

1. We restrict inference latency to close to the optimal latency ($20\%$). In this setting, the tight latency constraints make it impossible to batch the inference query, essentially reducing the search space for the optimal configuration.

2. In the second benchmark, we relax the inference latency constraint to include the configurations that provide the lowest energy-per-query in Figure~\ref{fig:pareto}. In this setting, we need to explore the batch size dimension to find the configuration that minimizes energy. We test on EfficientNet B0, B4, and B7, as well as Bert Base on both Jetson TX2 and Jetson Orin.

\textbf{Results: } 
Figure \ref{fig:bo_vs_rs} shows that CBO outperforms RS in both scenarios. Since CBO models the relationship between hardware configuration and latency, it can find a near-optimal solution with only $5$ to $15$ samples. In the second scenario, the performance of RS deteriorates as it is unable to leverage the relationship between latency and batch size when dealing with a multiplicatively increasing search space. Overall CBO takes 3-10x fewer samples in the second setting. The overhead of performing CBO is also minimal. As shown in Figure ~\ref{fig:bo_vs_rs}, CBO only requires around 15 samples to find a near-optimal solution and the optimization procedure can be completed in a few minutes. In cases where a new model is deployed, only a few minutes of overhead are needed to find optimal configurations for the new model.

It is important to note that though RS might achieve performance comparable to CBO under certain conditions, this result is merely the expected value and the variance of RS is large. For instance, if 10 out of 200 configurations are near-optimal, the expected number of trials needed to reach a near-optimal configuration is 20, with a standard deviation of 19.49. Consequently, it's plausible that even after 40 trials, RS might still fail to identify a near-optimal configuration. On the other hand, the standard deviation of CBO is smaller; in all experiments, CBO's standard deviations are less than 3.

\begin{table}[t]
 \centering
 \caption{This table shows the SLO violation rate and job completion time of various scheduling strategies for \textbf{Jetson Orin} on \textbf{EfficientNetB7}. The baseline shows the SLO violation rate without fine-tuning.}
\begin{tabular}{ llc } 

 \toprule
 \rowcolor{Gray}  Method & Workload & SLO violation\\
 \midrule
 Greedy & Uniform & 37.91\% \\ 
  \rowcolor{cornsilk} Adaptive & Uniform & \textbf{2.08\%} \\ 
 Greedy & Poisson & 60.41\% \\ 
  \rowcolor{cornsilk} Adaptive & Poisson & \textbf{5.42\%} \\ 
  Greedy & Twitter & 22.0\% \\ 
  \rowcolor{cornsilk} Adaptive & Twitter & \textbf{5.8\%} \\ 
  \midrule
 Baseline & Uniform & 0.4\% \\
 Baseline & Poisson & 1.67\% \\
 Baseline & Twitter & 3.8\% \\
  \bottomrule
 \end{tabular}
 \label{tab:schedulingOrin}
\end{table}

\begin{table}[t]
 \centering
 \caption{This table shows the SLO violation rate and job completion time of various scheduling strategies for \textbf{Jetson TX2} on \textbf{EfficientNetB4}. The baseline shows the SLO violation rate without fine-tuning.}
\begin{tabular}{ llc } 

 \toprule
 \rowcolor{Gray}  Method & Workload & SLO violation\\
 \midrule
 Greedy & Uniform & 16\% \\ 
\rowcolor{cornsilk} Adaptive & Uniform & \textbf{5.5\%} \\ 
 Greedy & Poisson & 35.4\% \\ 
\rowcolor{cornsilk} Adaptive & Poisson & \textbf{7.4\%} \\ 
Greedy & Twitter & 35.0\% \\ 
\rowcolor{cornsilk} Adaptive & Twitter & \textbf{7.5\%} \\ 
  \midrule
 Inference Only & Uniform & 1\% \\
 Inference Only & Poisson & 3.5\% \\
 Inference Only & Twitter & 5.3\% \\
  \bottomrule
 \end{tabular}
 \label{tab:schedulingTX2}
\end{table}

\subsection{Workload-aware Fine-tuning Scheduling}

Next, we evaluate how well PolyThrottle handles fine-tuning requests alongside inference. The central question we aim to address is whether our performance predictor can effectively identify and adjust accordingly when the SLO requirement is at risk of being violated, and if reducing the inference batch size and trading throughput can satisfy the latency SLO. To simulate this scenario, we generate two distinct inference arrival patterns (Uniform and Poisson) and use the publicly available Twitter trace \cite{twitter} and compare our adaptive scheduling approach to greedy scheduling, where a fine-tuning request is scheduled as soon as it arrives. The three arrival patterns represent scenarios that are highly controlled and bursty, respectively. In this context, we contrast PolyThrottle's adaptive scheduling mechanism with the greedy scheduling approach to assess the efficacy of PolyThrottle in meeting the desired SLO requirement.

\textbf{Experiment Settings: } We evaluate on both synthetic and real workloads. For synthetic workloads, we generate a stream of inference requests using both Uniform and Poisson distributions. For real-world workload, we first uniformly sample a day of Twitter streaming traces and then compute the variance of requests during each minute. We then picked the segment with the highest variance to test PolyThrottle's capability in handling request bursts \cite{twitter, romero2021infaas}. On Jetson Orin, we replay the stream for 30 seconds and measure the SLO violation rate during the replay using EfficientNet B7. Since each burst only lasts for a few seconds, this suffices to capture many bursts in the workload. We find that running the experiment for longer durations produces similar results. We set the fine-tuning batch size to 64, the number of fine-tuning iterations to 10, SLO to 0.7s, the output dimension to 1000, and an average of 8 inference requests per second. On Jetson TX2, we do the same experiment on EfficientNet B4. Due to memory constraints, we perform the fine-tuning batch size to 8, SLO to 1s, the output dimension to 100, and an average of 4 inference requests per second. We select a less performative model on TX2 to meet a reasonable SLO target (under 1s). The number of fine-tuning iterations is chosen based on the duration of the replay. We then measure the energy costs when deploying PolyThrottle at the default and optimal hardware frequency, respectively, to measure how much energy we save during this period. The optimal hardware frequency is obtained from results in Section~\ref{sec:pareto}.

For greedy scheduling, we employ a standard drop policy \cite{crankshaw2017clipper, shen2019nexus}, whereby a request is dropped if it has already exceeded its deadline. In the adaptive setting, we use the predictor to determine whether to drop an inference request. We also replay the inference request stream without fine-tuning requests to serve as a baseline.

\textbf{Results: } Table~\ref{tab:schedulingOrin} and ~\ref{tab:schedulingTX2} show the SLO violation rates under various workloads and latency targets. The findings indicate that greedy scheduling may lead to significant SLO violations owing to the interference introduced by the fine-tuning workload. In contrast, PolyThrottle's adaptive scheduling mechanism demonstrates the ability to achieve low SLO violation rates by dynamically adjusting configurations. The baseline figures in the table represent SLO violation rates in the absence of interference from fine-tuning requests.

Inherent variance in neural network inference resulted in $1\%$ of SLO violations in the case of Uniform distribution. However, bursts in the Poisson distribution and the Twitter workload generated more SLO violations. PolyThrottle's adaptive scheduling mechanism significantly reduces the SLO violation rate, meeting the SLO requirements while concurrently handling fine-tuning requests. Nevertheless, in several instances, we were unable to achieve near-zero SLO violation rates. This limitation can be attributed to the granularity of scheduling as we process the current batch of requests over an extended timespan due to interference from the fine-tuning workload.

We also \textbf{reduce energy consumption} by $14\%$ on EfficientNet B7 on Jetson Orin and by $23\%$ on EfficientNet B4 on Jetson TX2 across the workloads. We show in Appendix \ref{appen:pred} how PolyThrottle reacts to changing SLOs when there are outstanding fine-tuning requests.

\vspace{-0.3cm}

\section{Conclusion}
In this work, we examine the unique characteristics of energy consumption in neural network inference, especially for edge devices. We identified unique tradeoffs and dimensions between energy consumption and inference latency SLOs and empirically demonstrated hidden components in optimizing energy consumption. We then propose an optimization framework that automatically and holistically tunes various hardware components to find a configuration aligned with the Pareto Frontier. We empirically verify the effectiveness and efficiency of PolyThrottle. PolyThrottle also adapts to the need for fine-tuning and proposes a simple performance prediction model to adaptively schedule fine-tuning requests while keeping the online inference workload under the inference latency SLO whenever possible. We hope our study sheds more light on the hidden dimension of NN energy optimization.

\bibliographystyle{ieeetr}
\bibliography{neurips_2023}



\newpage
\newpage
\appendix
\section{Hardware Details} \label{appen:measurement}
\subsection{Jetson platform details}
The \textbf{Jetson TX2 Developer Kit} features a 256-core NVIDIA Pascal GPU, a Dual-Core NVIDIA Denver 2 64-bit CPU, a Quad-Core ARM Cortex-A57 MPCore CPU, and 8GB of 128-bit LPDDR4 memory with 59.7 GB/s bandwidth. The kit's maximum power consumption is 15W. The \textbf{Jetson Orin Developer Kit} includes a 2,048-core NVIDIA Ampere GPU with 64 Tensor Cores and a 12-core Arm CPU. This kit comes with 32GB of 256-bit LPDDR5 memory, featuring a 204.8GB/s bandwidth, and has a maximum power consumption of 60W.

\subsection{Power consumption measurement} 
The Nvidia Jetson TX2 Developer Kit allows for separate measurements of GPU, CPU, DDR, and total energy consumption, while the Jetson Orin uses the built-in tegrastats module for measuring power usage across hardware components. Due to power rail design limitations, GPU power usage on the Jetson Orin can only be measured alongside SoC power usage. 

On Jetson TX2, we measure power usage by querying the total power input. We then average the peak power consumption to obtain the power usage during inference. Then we compute the energy cost for each inference request by multiplying the power and the inference time. On Jetson Orin, we leverage the existing tegrastats tool and repeatedly query tegrastats at a fixed interval (50ms). We then sum up each component's power consumption to obtain the overall power consumption, before multiplying the power and the inference time to obtain the energy cost for each inference request. To obtain a steady reading, we send 1000 inference requests for each hardware configuration for every model that we test. 

We cross-validate our measurements using a USB digital multimeter capable of transmitting data to computer software in real-time via Bluetooth. The measurements obtained from the multimeter generally align with those from the internal power rails on Jetson Kits, although external measurements are consistently around $10\%$ higher than Jetson internal measurements. This discrepancy may be attributable to unaccounted factors in the power rail design. We opted to use internal measurements since they provide component-specific readings, whereas the multimeter can only measure overall energy consumption. Moreover, the multimeter supports one measurement per second, while Jetson tools allow for millisecond-scale measurements, which are better suited to inference workloads.

\subsection{Measurement Overhead}
Since we repeatedly query the power input or built-in power management tool, we want to understand whether these queries affect total energy consumption. We use a USB digital multimeter capable of transmitting data to computer software in real-time via Bluetooth. We then run our inference program with and without querying the power input or the power management tool. We find that the power consumption reported by the multimeter increases around $5 \to 10\%$, depending on the base power consumption. We observe that this increment is near constant across different models and runs and therefore we believe using internal measurement as described in the section above will not affect our findings. The multimeter cannot provide the precision and flexibility we need to measure the energy cost of inference, which often operates at a millisecond scale. 

\section{Experimental Results} \label{appen:exp}

In this section, we further demonstrate the tradeoff between memory frequency and maximum GPU frequency by presenting an array of results. These results underline the interesting observation that the energy consumption patterns may vary for the same model operating on different devices. Furthermore, even for the same model-device pairing, the optimization landscape can be significantly influenced by the batch size. This underlines the complexities of energy optimization and the need for an adaptive framework that can take these factors into account. Figures $6 - 12$ show the energy consumption patterns of EfficientNet and Bert on Jetson TX2 and Orin under various batch sizes. Table \ref{tab:CPU} shows the optimal CPU frequency and corresponding energy consumption reduction in image preprocessing.

\begin{figure}[H]
\centering
\includegraphics[width=0.45\linewidth]{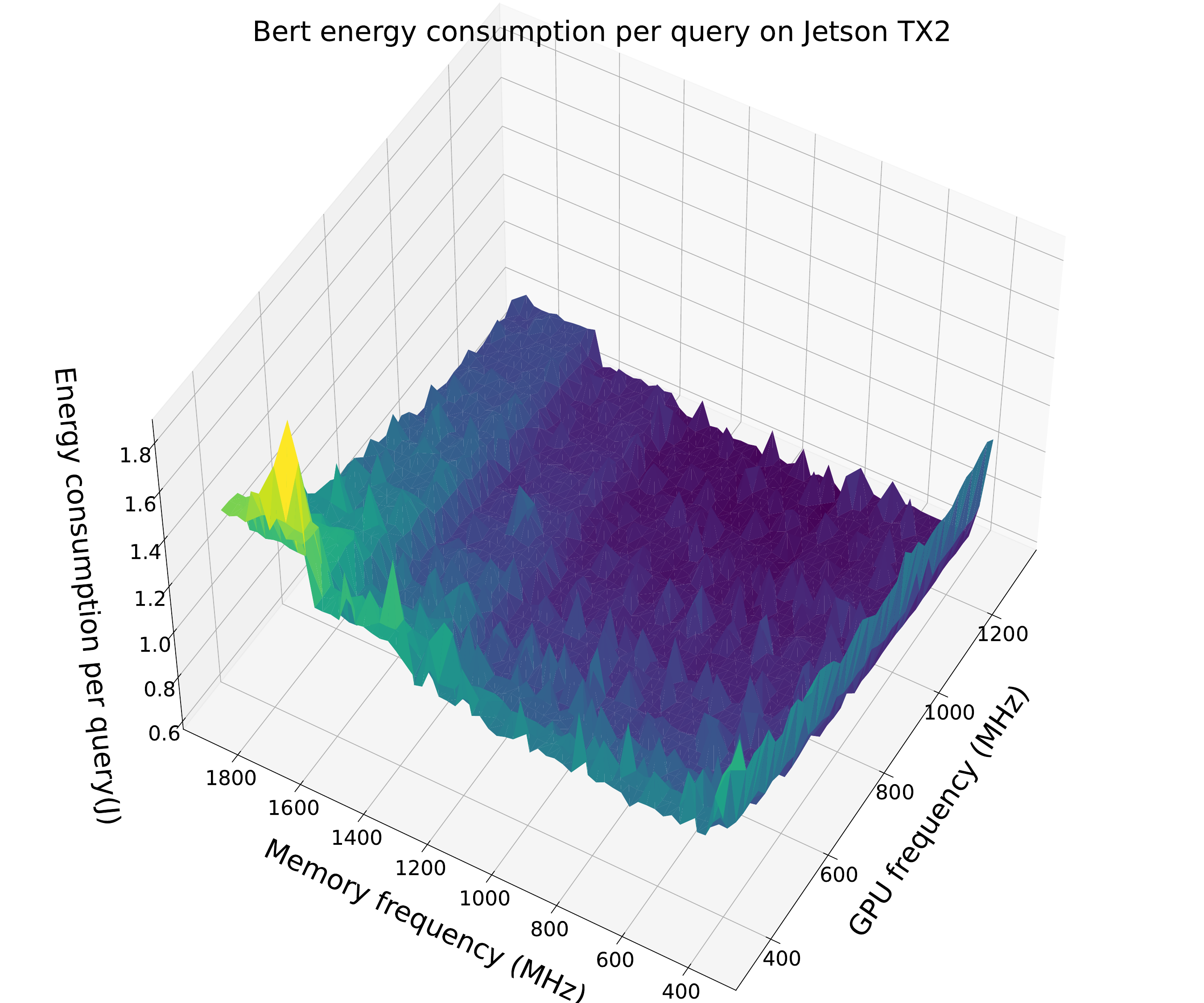}
\includegraphics[width=0.45\linewidth]{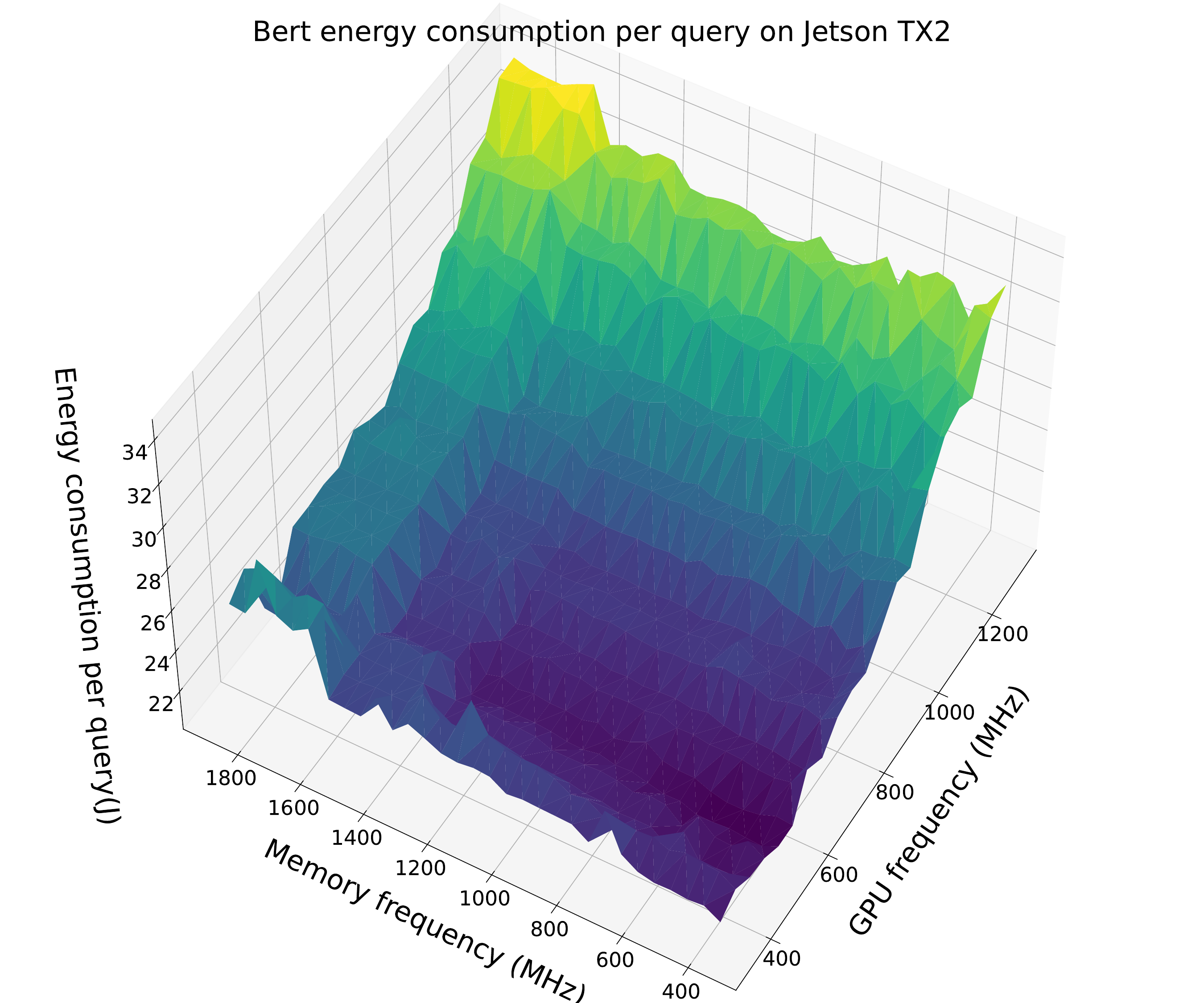}
\caption{Left figure shows per query energy cost as we vary the GPU frequency and memory frequency for Bert at FP16 on Jetson TX2 versus varying Memory and GPU frequency with batch size fixed at 1. Right figure shows per query energy cost as we vary the GPU frequency and memory frequency for Bert at FP32 on Jetson TX2 versus varying Memory and GPU frequency with batch size fixed at 1.}
\label{fig:3d-bert-tx2}
\label{fig:3d-bert-fp32-tx2}
\end{figure}

\begin{figure}[H]
\centering
\includegraphics[width=0.45\linewidth]{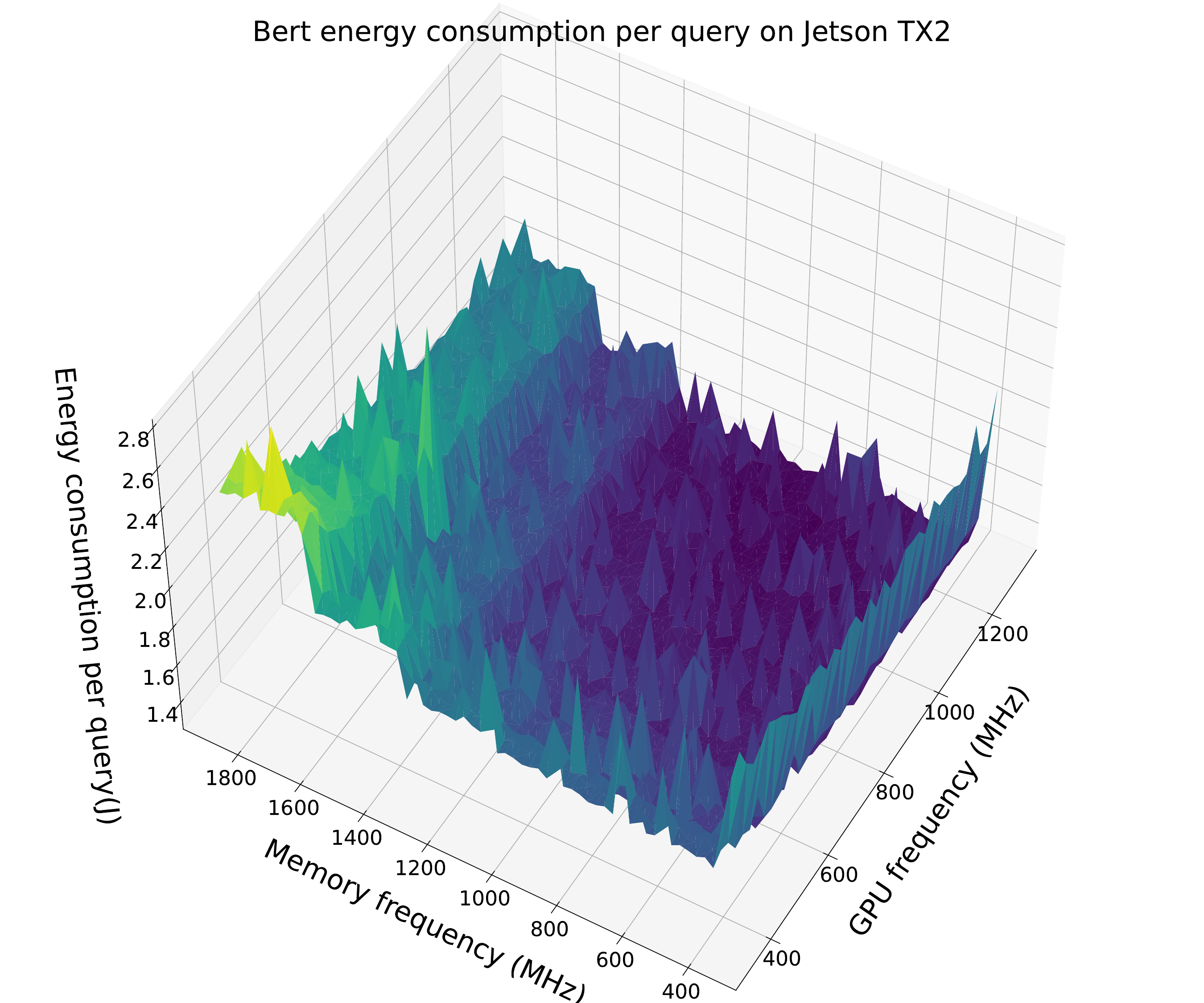}
\includegraphics[width=0.45\linewidth]{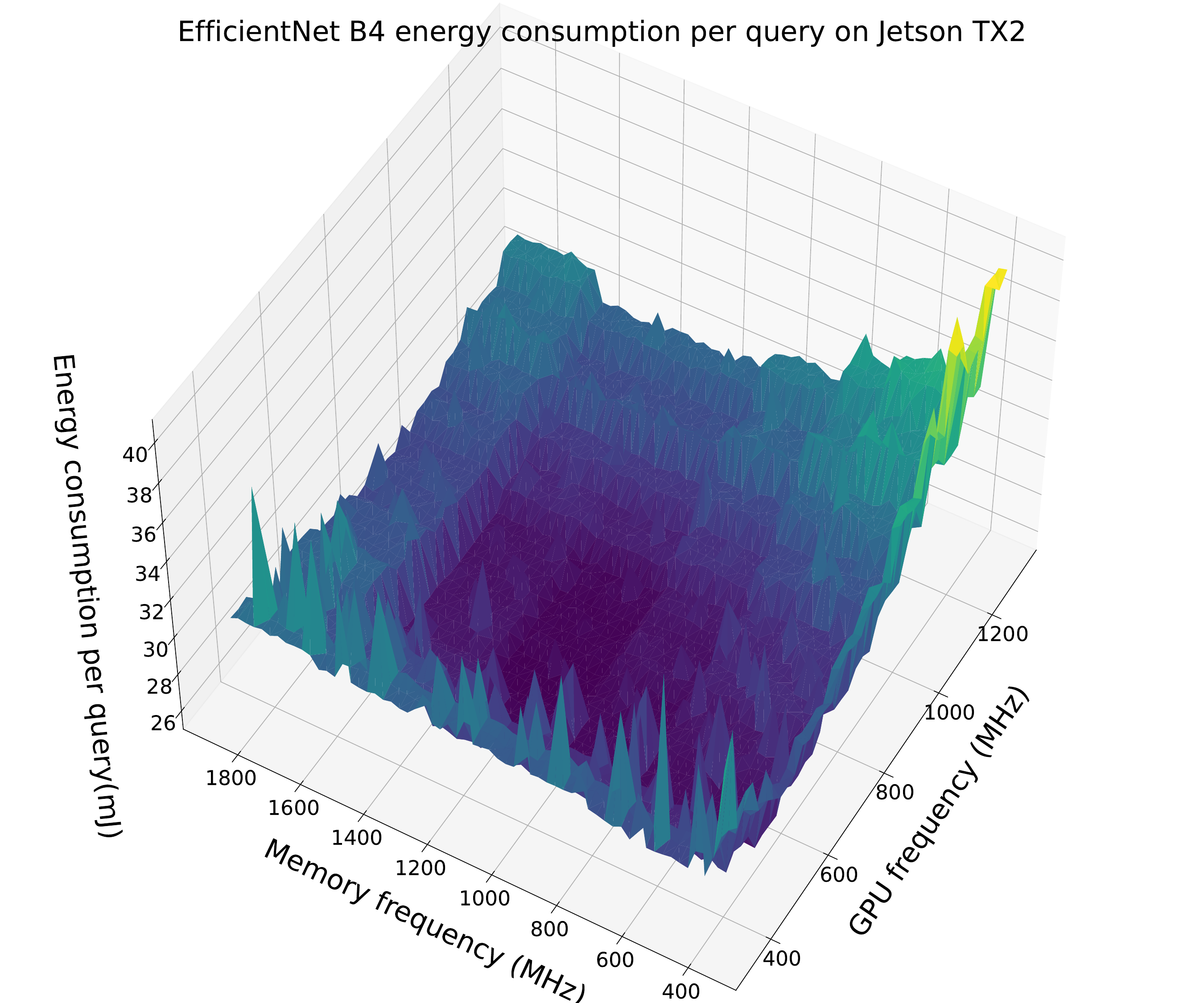}
\caption{Left figure shows per query energy cost as we vary the GPU frequency and memory frequency for Bert at FP16 on Jetson TX2 versus varying Memory and GPU frequency with batch size fixed at 8. Right figure shows per query energy cost as we vary the GPU frequency and memory frequency for EfficientNet B4 at FP16 on Jetson TX2 versus varying Memory and GPU frequency with batch size fixed at 16.}
\label{fig:3d-bert-fp16-b8-tx2}
\label{fig:3d-enetb4-fp16-b8-tx2}
\end{figure}

\begin{figure}[H]
\centering
\includegraphics[width=0.45\linewidth]{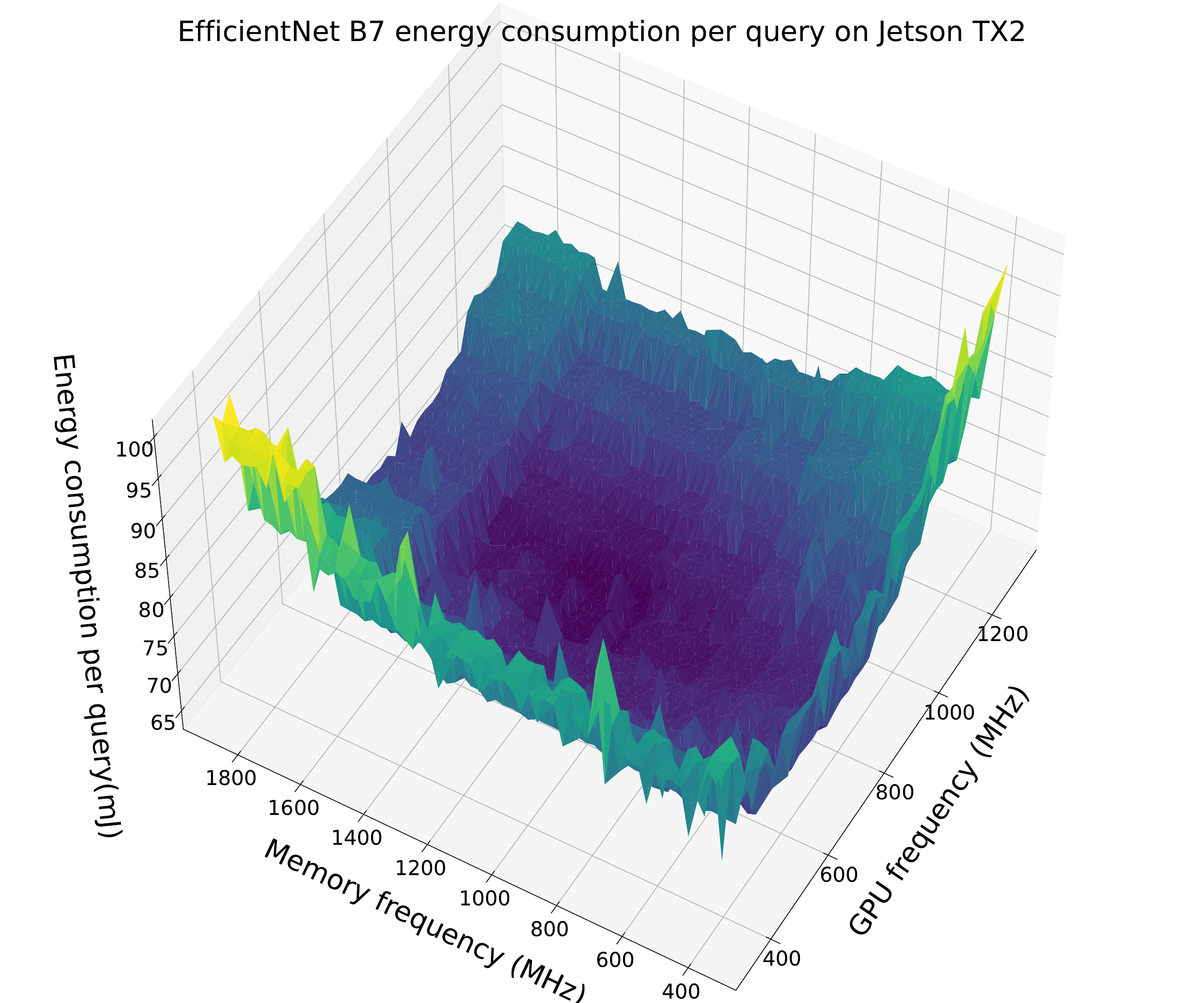}
\includegraphics[width=0.45\linewidth]{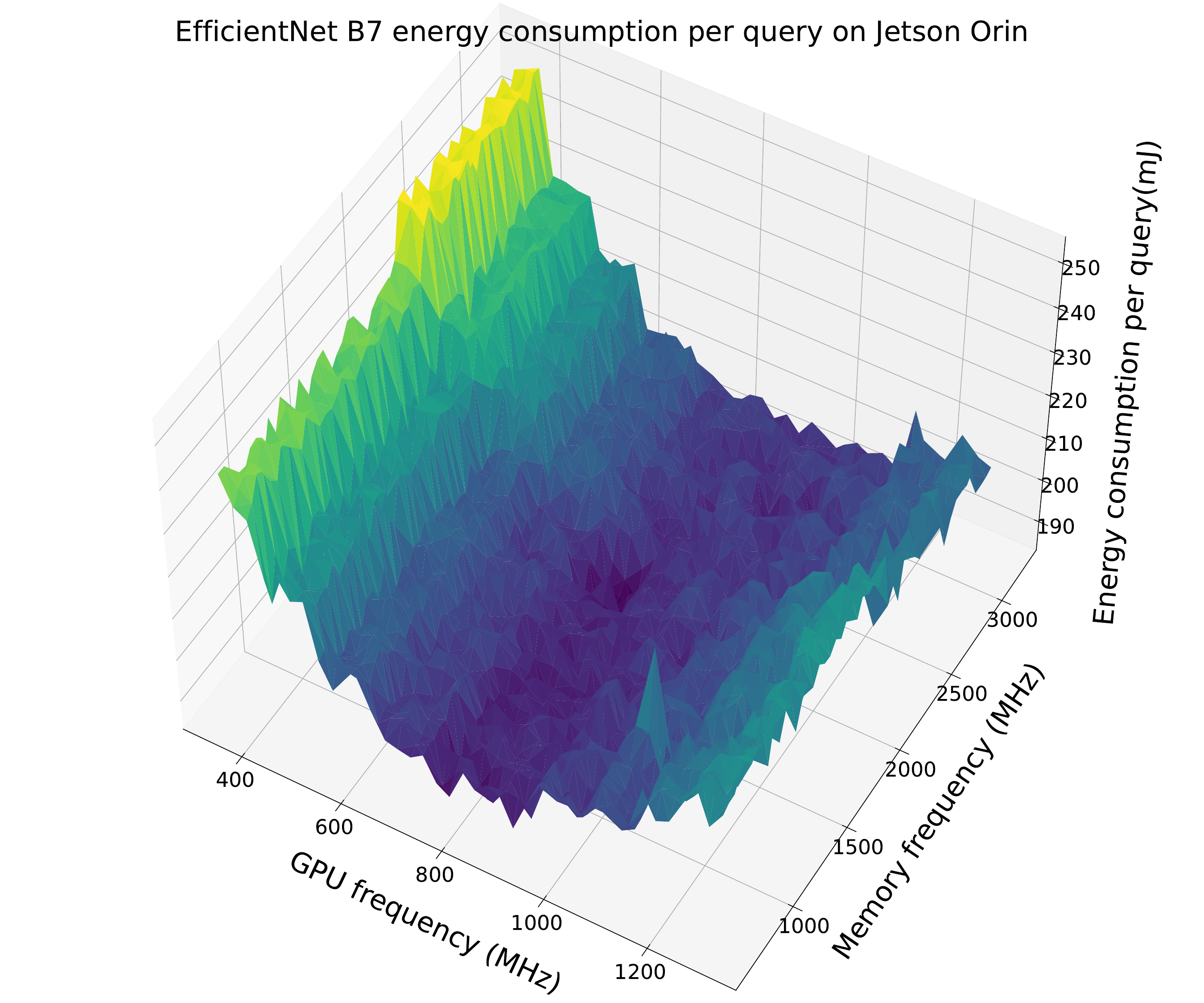}
\caption{Left figure shows per query energy cost as we vary the GPU frequency and memory frequency for EfficientNet B7 at FP16 on Jetson TX2 versus varying Memory and GPU frequency with batch size fixed at 16. Right figure shows per query energy cost as we vary the GPU frequency and memory frequency for EfficientNet B7 at FP16 on Jetson Orin versus varying Memory and GPU frequency with batch size fixed at 8.}
\label{fig:3d-enetb7-fp16-b8-tx2}
\label{fig:3d-enetb7-fp16-b8-orin}
\end{figure}

\begin{figure}[H]
\centering
\includegraphics[width=0.45\linewidth]{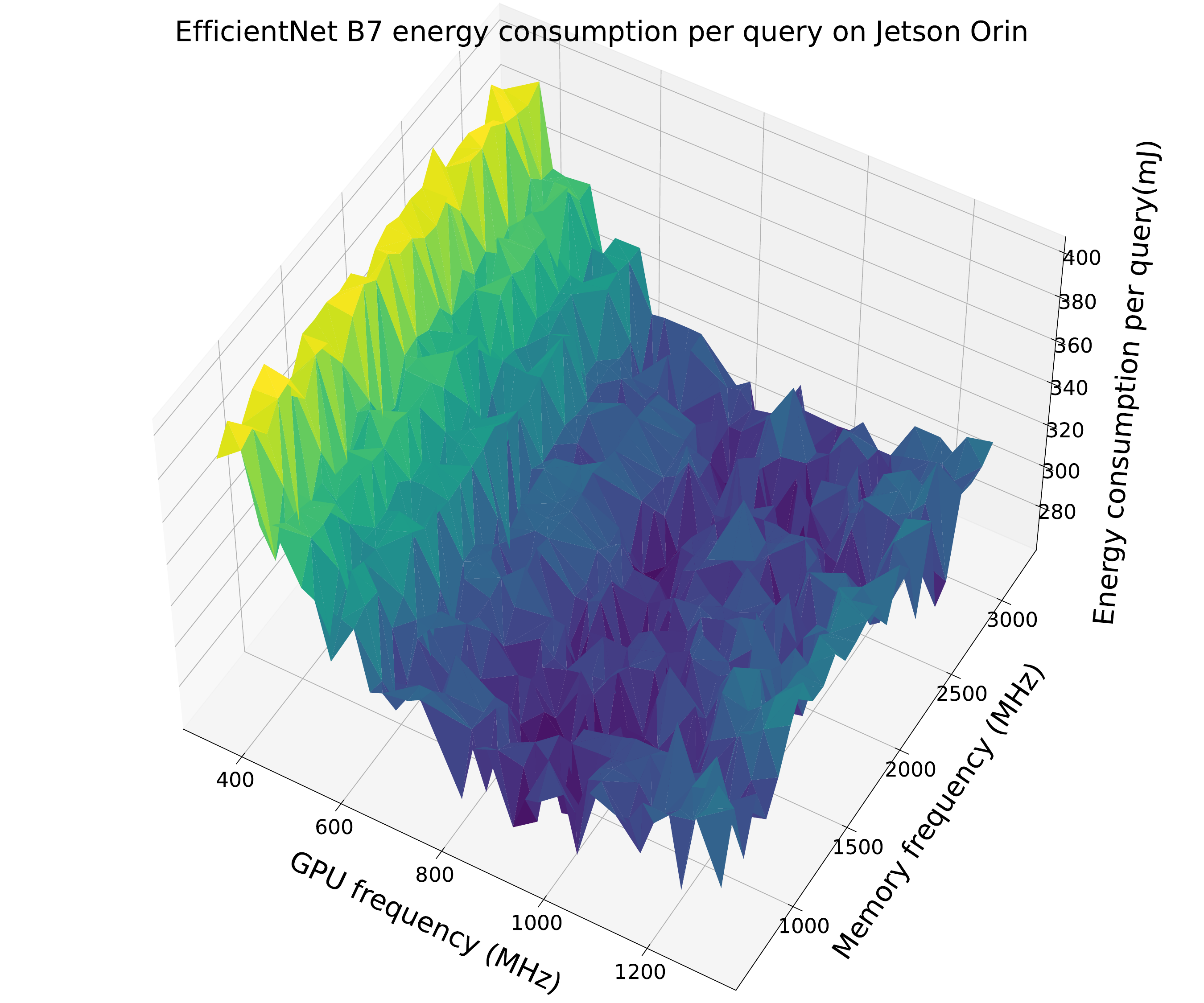}
\includegraphics[width=0.45\linewidth]{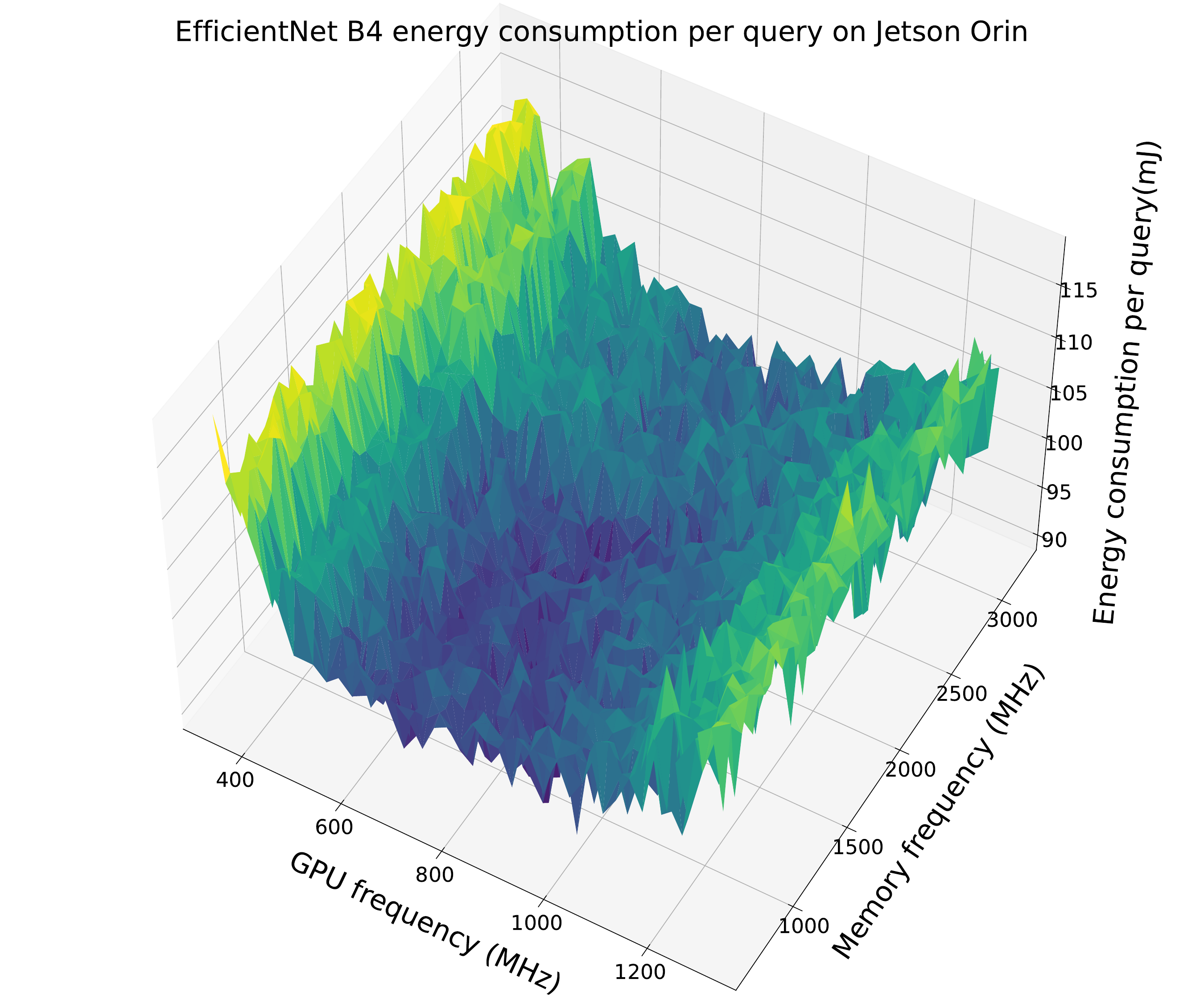}
\caption{Left figure shows per query energy cost as we vary the GPU frequency and memory frequency for EfficientNet B7 at FP16 on Jetson Orin versus varying Memory and GPU frequency with batch size fixed at 1. Right figure shows per query energy cost as we vary the GPU frequency and memory frequency for EfficientNet B4 at FP16 on Jetson Orin versus varying Memory and GPU frequency with batch size fixed at 8.}
\label{fig:3d-enetb7-fp16-b1-orin}
\label{fig:3d-enetb4-fp16-b8-orin}
\end{figure}

\begin{table}[t]
 \centering
 \caption{This table shows how CPU frequency (MHz) affects energy consumption for image processing on Jetson TX2 and Jetson Orin.}
\begin{tabular}{ lccc } 

 \toprule
\rowcolor{Gray}  &  & Optimal Min \\
\rowcolor{Gray} Device & \multirow{-2}{*}{{Energy Reduction}}& CPU Freq\\
 \midrule
 Orin & 28.6\% & 900\\ 
 TX2 & 29.4\% & 1000\\ 
  \bottomrule
 \end{tabular}
 \label{tab:CPU}
\end{table}

\section{Arithmetic intensity}
The arithmetic intensity of a 2D convolution layer can be computed as:

\begin{align}
AI_{conv} = \frac{N \cdot K \cdot P \cdot Q \cdot C \cdot R \cdot S}{ N \cdot C \cdot H \cdot W + K \cdot C \cdot R \cdot S + N \cdot K \cdot P \cdot Q}
\label{eq:AIConv}
\end{align}

The notations used in equation \ref{eq:AIConv} can be found in table \ref{tab:notation}.

\begin{table}[t]
 \centering
 \caption{The notation table defines the variable used in equation \ref{eq:AIConv}}
\begin{tabular}{ cl } 

 \toprule
 \rowcolor{Gray} Variable & Definition \\
 \midrule
 $N$ & Batch size \\ 
 $C$ & Input number of channels \\ 
 $H$ & Input tensor height \\
 $W$ & Input tensor width \\ 
 $K$ & Output number of channels \\ 
 $P$ & Output tensor height  \\
 $Q$ & Output tensor width \\ 
 $R$ & Convolution filter height\\ 
 $S$ & Convolution filter width \\
 
  \bottomrule
 \end{tabular}
 \label{tab:notation}
\end{table}

The FLOPs term captures the total computation of each workload, while the arithmetic intensity term captures how much computation power and memory bandwidth will affect the final performance. Combining the aforementioned features with an intercept term, which captures the fixed overhead in neural network inference, we can build a model that predicts inference latency if the hardware operating frequency is stable.

\section{Predictor Analysis} \label{appen:pred}
\begin{figure}
\centering
\includegraphics[width=0.5\linewidth]{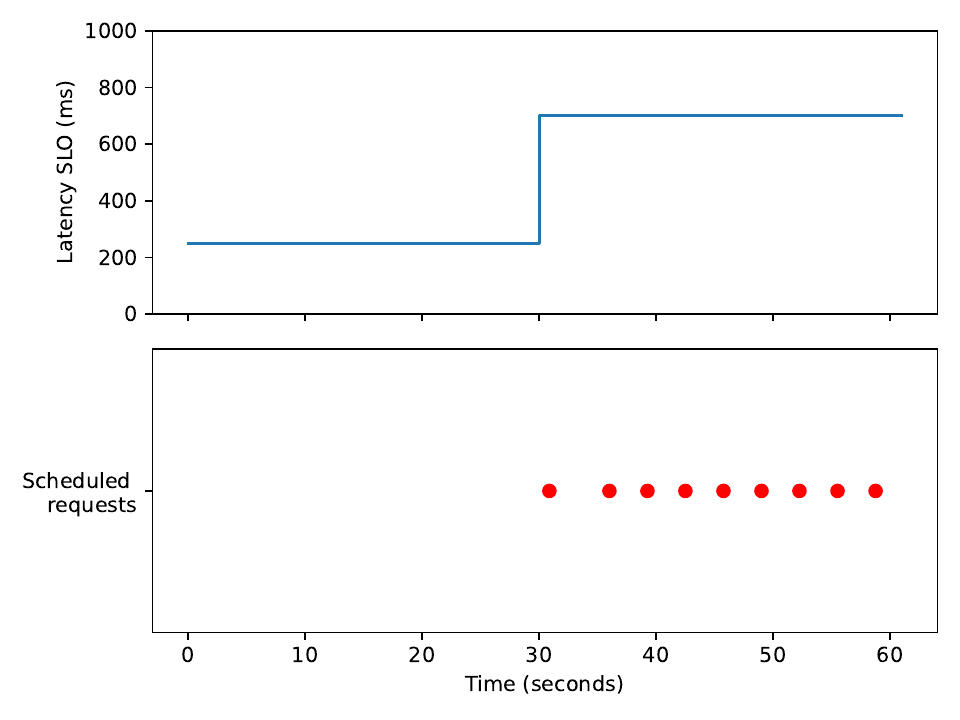}
\caption{This figure shows the inference latency SLO and when fine-tuning requests are scheduled.}
\label{fig:ft-scheduling.pdf}
\end{figure}

We vary the latency SLO to assess how the predictor schedules the fine-tuning requests. We replay a 60-second stream where we initially set the latency SLO to 250ms for the first half (30 seconds), and then increase it to 700ms for the remainder. As shown in Figure \ref{fig:ft-scheduling.pdf}, under stringent latency conditions, the predictor deduces that it is impractical to schedule fine-tuning requests while adhering to the latency SLO, hence no fine-tuning requests are scheduled. Conversely, when the latency SLO is more relaxed, the predictor determines that it is feasible to schedule fine-tuning requests and sequentially schedules each request once the preceding one is completed and has issued a completion signal.

\end{document}